\documentclass[10pt,journal]{IEEEtran}

\usepackage{cite}
\usepackage{amsmath,amssymb,amsfonts}
\usepackage{graphicx}
\usepackage{textcomp}
\usepackage{xcolor}
\def\BibTeX{{\rm B\kern-.05em{\sc i\kern-.025em b}\kern-.08em
		T\kern-.1667em\lower.7ex\hbox{E}\kern-.125emX}}

\usepackage{bm}
\usepackage{hhline}
\usepackage{wasysym}
\usepackage{amssymb}
\usepackage{stfloats}
\usepackage{xspace}
\usepackage{makecell}
\usepackage{colortbl}
\usepackage{arydshln}
\usepackage{multirow}
\usepackage{multicol}
\usepackage{threeparttable}
\usepackage[switch]{lineno}
\usepackage{url}

\usepackage{breakurl}
\usepackage{pifont}
\makeatletter
\newif\if@restonecol
\makeatother

\usepackage[linesnumbered,ruled,vlined]{algorithm2e}
\usepackage{algpseudocode}
\usepackage{multirow}
\usepackage{enumerate}
\usepackage[normalem]{ulem}
\usepackage{enumitem}
\newlist{steps}{enumerate}{1}
\setlist[steps, 1]{label = Step \arabic*:}
\usepackage{float}
\usepackage{listings}
\usepackage{color}
\definecolor{codegreen}{rgb}{0,0.6,0}
\definecolor{codegray}{rgb}{0.5,0.5,0.5}
\definecolor{codepurple}{rgb}{0.58,0,0.82}
\definecolor{backcolour}{rgb}{0.95,0.95,0.92}
\lstdefinestyle{mystyle}{
	commentstyle=\color{codegreen},
	keywordstyle=\color{magenta},
	numberstyle=\tiny\color{codegray},
	stringstyle=\color{codepurple},
	basicstyle=\footnotesize\ttfamily,
	breakatwhitespace=false,
	breaklines=true,
	captionpos=b,
	keepspaces=true,
	numbers=left,
	numbersep=5pt,
	showspaces=false,
	showstringspaces=false,
	showtabs=false,
	tabsize=2,
	captionpos=b,
	abovecaptionskip=\smallskipamount,
	frame=lines,
}

\newcommand{\FuPruner}{\textsc{FuPruner}\xspace}

\usepackage{fancyhdr}

\begin{document}

\title{Fusion-Catalyzed Pruning for Optimizing Deep Learning on Intelligent Edge Devices}

\author{Guangli~Li,
        Xiu~Ma,
        Xueying~Wang,
        Lei~Liu,
        Jingling~Xue,
        and~Xiaobing~Feng
\thanks{This work is supported by the National Key R\&D Program of China (2017YFB1003103), the Science Fund for Creative Research Groups of the National Natural Science Foundation of China (61521092), and the Australian Research Council Grants (DP170103956 and DP180104069). (\textit{Corresponding
author: Lei Liu.})}
\thanks{G.~Li, X.~Wang, L.~Liu, and X.~Feng are with
the State Key Laboratory of Computer Architecture, Institute of Computing Technology, Chinese Academy of Sciences, Beijing 100190, China
, and also with the University of Chinese Academy of Sciences, Beijing 100190, China (email: liguangli@ict.ac.cn, wangxueying@ict.ac.cn, liulei@ict.ac.cn, fxb@ict.ac.cn).}
\thanks{X.~Ma is with the College of Computer Science and Technology, Jilin University, Changchun 130012, China (email: maxiu18@mails.jlu.edu.cn).}
\thanks{J.~Xue is with the School of Computer Science and Engineering, University of New South Wales, Sydney, NSW 2052, Australia (email: jingling@cse.unsw.edu.au).}
}

\maketitle

\thispagestyle{fancy}
\fancyhead{}
\lhead{\footnotesize \copyright~2020 IEEE.  Personal use of this material is permitted.  Permission from IEEE must be obtained for all other uses, in any current or future media, including reprinting/republishing this material for advertising or promotional purposes, creating new collective works, for resale or redistribution to servers or lists, or reuse of any copyrighted component of this work in other works. The final version of record is available at \url{http://dx.doi.org/10.1109/TCAD.2020.3013050}}
\lfoot{}
\cfoot{}
\rfoot{}

\begin{abstract}
The increasing computational cost of deep neural network models limits the applicability of intelligent applications on resource-constrained edge devices.
While a number of neural network pruning methods have been proposed to compress the models, prevailing approaches focus only on parametric operators (e.g., convolution), which may miss optimization opportunities.
In this paper, we present a novel fusion-catalyzed pruning approach, called \FuPruner, which simultaneously optimizes the parametric and non-parametric operators for accelerating neural networks.
We introduce an aggressive fusion method to equivalently transform a model, which extends the optimization space of pruning and enables non-parametric operators to be pruned in a similar manner as parametric operators,
and a dynamic filter pruning method is applied to decrease the computational cost of models while retaining the accuracy requirement.
Moreover, \FuPruner provides configurable optimization options for controlling fusion and pruning, allowing much more flexible performance-accuracy trade-offs to be made.
Evaluation with state-of-the-art residual neural networks on five representative intelligent edge platforms, Jetson TX2, Jetson Nano, Edge TPU, NCS, and NCS2, demonstrates the effectiveness of our approach, which can accelerate the inference of models on CIFAR-10 and ImageNet datasets.
\end{abstract}


\section{Introduction}
\label{sec:intro}

\IEEEPARstart{D}{eep} neural networks (DNNs) have achieved remarkable performance in various intelligence tasks, such as object recognition~\cite{he2016deep},
and become increasingly popular in mobile and embedded platforms~\cite{li2018learning}.
However, the inference of neural networks, i.e., obtaining the predicted result by a pre-trained model with a given input is a computation-intensive task, performed cumbersomely on edge devices, which is limited by the tight resource constraints, including processor, energy and memory.
Despite the fact that the inference procedure can be performed in the cloud,
i.e., computation of DNNs can be offloaded to high-performance devices (e.g., cloud servers),
this approach is not suitable in several scenarios due to the security concerns, real-time constraints, and unstable network connectivity.
As such, the technique for supporting on-device inference has sparked an interest in both industry and academia~\cite{cheng2018recent,hadidi2019characterizing}.
On the one hand, several optimization methods for compressing and accelerating neural networks, such as pruning~\cite{blalock2020state}, have been proposed, which bring tolerable accuracy loss but decrease the cost of computation and storage.
On the other hand, developing intelligent edge devices equipped with specialized hardware, such as Google Tensor Processing Unit (TPU)~\cite{tpu} and Intel Neural Computer Stick (NCS)~\cite{ncs}, becomes an inevitable trend due to the inefficiency of general-purpose hardware platforms (e.g., CPU) for executing intelligent applications.

\begin{figure}[t]
	\centering
	\includegraphics[width=0.95\linewidth]{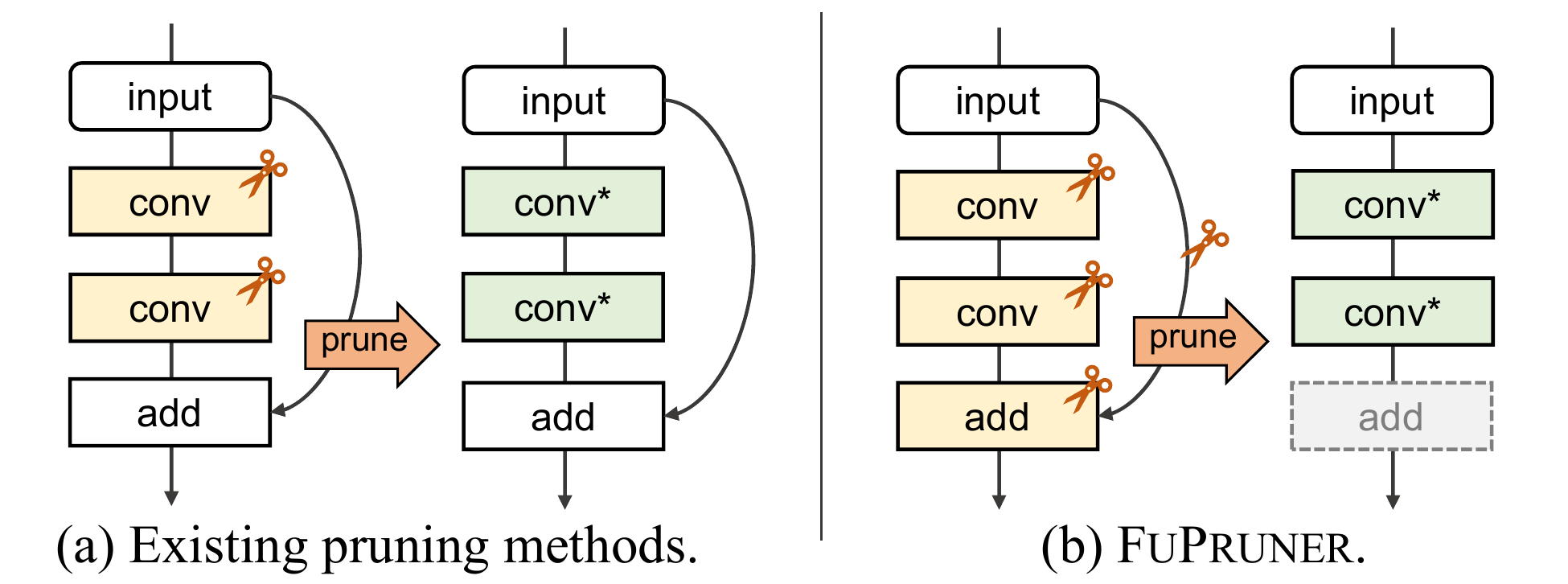}
	\caption{Comparing optimized neural network operators in existing pruning methods with \FuPruner. The operator to be optimized is marked as the yellow box while the pruned operator is marked as the green box. The pruned operator reduces the number of parameters compared with the original one and is marked with an ``$*$''. Especially, the element-wise addition operator has been removed after pruning in \FuPruner, as marked by the dashed gray~box.}
	\label{fig:compare_methods}
\end{figure}

Neural network pruning is one of the effective ways to optimize models by reducing redundant neurons and connections.
Prior works on weight pruning~\cite{han2015learning,guo2016dynamic,dong2018fast} have succeeded in achieving high sparsity as well as theoretically attractive speedups.
However, the non-structured schemes (i.e., irregular sparsity) of pruned DNN operators are not implementation friendly, which can hardly reach expected acceleration in real-world applications.
In contrast, filter pruning methods~\cite{luo2017thinet,he2017channel,he2018soft}, which selectively reduce unimportant filters, shrink a model into a thinner one and achieve structured sparsity.
As such, the filter-level pruned models can adequately enjoy the benefits of decreased convolution filters, providing realistic performance improvements.
For intelligent edge devices, their run-time systems are usually closed-source or unmodifiable, limiting the applicability of framework-dependent optimizations  such as low-precision quantization~\cite{guo2018survey}.
In addition, some of these optimizations also introduce extra computation overheads.
In contrast, the models optimized by filter pruning will be independent from specific frameworks or libraries, exhibiting
better applicability and generality.

Nevertheless, the existing filter pruning methods~\cite{luo2017thinet,he2017channel,he2018soft} focus only on optimizing models by pruning parametric operators such as convolution, as shown in Figure~\ref{fig:compare_methods}.
The non-parametric operators like an element-wise operation are generally considered to be unimportant on the general-purpose platforms such that they are rarely considered by neural network pruning optimization, whereas it is a different story on specialized intelligence platforms.
In this paper, we analyze the fine-gained operator-level performance of DNN models, rather than end-to-end performance in previous studies~\cite{hadidi2019characterizing},
and find that a non-parametric operator also plays an important role when executed on the intelligent edge devices due to its time-consuming process (Section~\ref{sec:motivation}), which has, however, been neglected before.
To accelerate DNN models on intelligent edge devices effectively, it is thus imperative to take the characteristics of specialized intelligence hardware into account.

In this paper, we present \FuPruner (\underline{Fu}sion-catalyzed \underline{Pruner)}, a novel optimization approach that, unlike prior work, prunes non-parametric and parametric operators simultaneously, aiming to accelerate the neural network inference on intelligent edge devices with as little accuracy loss as possible.
The key idea is a new aggressive fusion scheme for enlarging the optimization space for pruning, by transforming existing operators and
inserting auxiliary ones while preserving the equivalence of neural network models, so that non-parametric operators can be pruned in the same manner as parametric operators.
Moreover, \FuPruner supports configurable optimizations, including fusion option and pruning rate to enable flexible trade-offs between the accuracy loss and inference performance, making it possible to optimize models while meeting the accuracy requirement.
To the best of our knowledge, \FuPruner is the first neural network pruning approach with the ability to prune non-parametric operators.

In summary, this paper makes the following contributions:

\begin{itemize}		
	\item We propose a novel fusion-catalyzed pruning approach, namely \FuPruner, to accelerate deep neural networks on intelligent edge devices by pruning parametric and non-parametric operators simultaneously.
	
	\item Our pruning technique facilitates flexible optimization configurations, which allows users to optimize a model in accordance with its requirements, thereby realizing reasonable accuracy-performance trade-offs.

	\item We demonstrate the effectiveness and efficiency of \FuPruner by optimizing state-of-the-art deep neural networks on CIFAR-10 and ImageNet datasets.
	
\end{itemize}

\section{Background and Related Work}

In recent years, edge intelligence\cite{li2018learning} has become more and more popular, which benefits from the unprecedented success of deep learning, making the research on efficient on-device inference becomes an inevitable trend.
In this section, we first introduce the existing optimization technologies for accelerating DNN models from the perspectives of software down to hardware and then review the work most closely related to this paper, namely, neural network pruning (Section \ref{sec::related_pruning}) and operator fusion (Section \ref{sec::related_fusion}).

\begin{itemize}[leftmargin=2.5ex]
	\item \textbf{Model Compression.}
	Recent studies on model compression techniques, including neural network pruning~\cite{blalock2020state} and low-precision quantization~\cite{guo2018survey}, modify  neural networks by reducing redundancy, thereby realizing inference acceleration.
	In general, re-training processes are required for maintaining the accuracy of optimized models.
	The optimized model parameters are reconstructed from the original ones, and  the model compression can thus be regarded as an optimization technique in the model design and development stage.
	
	\item \textbf{Inference Frameworks.}
	Existing deep learning frameworks for on-device inference, including TensorFlow Lite \cite{tf-lite}, PyTorch Mobile~\cite{pytorch-mobile}, NCNN~\cite{ncnn}, and MNN~\cite{jiang2020mnn}, are equipped with the ability to analyze and optimize the trained models by using system-level optimization (e.g., data layout selection, operator fusion and parallelization) in the deployment stage.
	The major difference between the optimizations in these inference frameworks and the model compression is that the former usually focus only on implementation without modifying the model itself.
	
	\item \textbf{Specific Hardware.}
	To solve the problem of inefficiency on general-purpose hardware platform (e.g., CPU), many intelligent edge devices with specialized hardware exist, including Google Edge TPU~\cite{tpu}, Intel NCS~\cite{ncs}, NVIDIA Jetson TX2~\cite{tx2}, and NVIDIA Jetson Nano~\cite{nano}.
	In addition, several studies, such as VTA~\cite{moreau2018vta}, deployed custom-hardware designs based on FPGAs.
	While DNN models can be deployed on these inteligent edge platforms by specific runtime systems, which are usually closed-source or unmodifiable, the model compression techniques can be used to further optimize the inference performance.
	
\end{itemize}

Besides, there are several studies on the adaptive inference for optimizing deep learning on embedded platforms, including adaptive strategies for neural network inference~\cite{panda2016conditional, DBLP:conf/iclr/YuYXYH19, panda2017falcon, marco2020optimizing} and \mbox{hardware/software} co-design~\cite{tann2016runtime, jayakodi2018trading, jayakodi2020design}, which allow deep neural networks to be configurable and executed dynamically
at runtime based on the resource constraints.

Our approach represents the first for model compression, which
essentially
optimizes the DNN model by pruning and re-training.
Especially, we have designed an aggressive operator fusion scheme to solve new problems in pruning domains and our proposed approach is applied for accelerating models on some specific hardware finally.
The proposed approach has two major advantages over the prior work. First,
\FuPruner represents a framework-independent approach,
increasing
its applicability in practice, as it
requires
neither special implementations nor run-time system modifications. Second,
\FuPruner does not incur any extra run-time overhead,
make it well suited to resource-constrained edge platforms.

\subsection{Neural Network Pruning Methods}
\label{sec::related_pruning}
Unlike the early efforts on weight pruning methods~\cite{han2015learning,guo2016dynamic,dong2018fast} that may cause the unstructured sparsity,
filter pruning methods shrink a DNN model into a thinner one by reducing the redundant convolutional filters or channels, making structured sparsity for optimized models to achieve realistic acceleration.
Li \emph{et~al.} \cite{li2016pruning} pruned unimportant filters with small $\ell_1$-norm.
He \emph{et~al.} \cite{he2017channel} performed a channel-level pruning based on LASSO regression and least square reconstruction.
Lou \emph{et~al.} \cite{luo2017thinet} selected the filter to be pruned according to the statistics information.
In addition to above studies on pruning filters directly in an unrecoverable manner, recently, He \emph{et~al.} \cite{he2018soft,he2019asymptotic} proposed a ``soft pruning manner'' by dynamically pruning filters during training, which enlarges the optimization space and remains the model capacity, enabling the pruned filters to be potentially recovered, thereby realizing higher accuracy.
Although the pruning methods are advocated in accelerating models, the existing approaches focus only on parametric operators (e.g., convolution).
In this paper, our proposed approach can prune non-parametric and parametric operators simultaneously, leading to more efficiency for neural network inference on intelligent edge devices.
To the best of our knowledge, \FuPruner is the first pruning approach with the ability of optimizing non-parametric operators.

\subsection{Operator Fusion Techniques}
\label{sec::related_fusion}
In general, deep learning frameworks represent a neural network architecture as a graph-based intermediate representation (IR)~\cite{cyphers2018intel,rotem2018glow,dong2019acorns}, known as a computation graph.
Operator fusion is a commonly used optimization~\cite{li2020deep}, which fuses two or more adjacent operators into a larger operator with coarser granularity so that the overheads can be reduced as well as the further low-level optimization can be facilitated.
For example, it can eliminate the intermediate results incurred by executing a lot of fine-grained operators, reducing the overheads of data transformation and kernel invocation.
Current deep learning frameworks, including TensorRT~\cite{vanholder2016efficient}, TensorFlow~\cite{abadi2016tensorflow}, MXNet~\cite{chen2015mxnet}, and TVM~\cite{chen2018tvm}, perform operator fusion for a computation graph by using rule-based strategies repetitively, by reducing gradually the number of operators to be performed.
Besides, Jia \emph{et~al.} proposed MetaFlow~\cite{jia2019optimizing} and TASO~\cite{jia2019taso} to further explore the optimization space on graph substitutions.
The fusion rules are formally verified in TASO, thereby achieving not only efficient but also correct model optimizations at the system level.
Currently, most of the operator fusion techniques in the inference framework are performed in the deployment stage, i.e., they are used to optimize the models compressed in the development stage.
In this paper, we propose a new operator fusion scheme to improve the pruning process from the perspective of model compression, which enables \FuPruner to optimize non-parametric operators similarly as parametric operators.

\section{Motivation}
\label{sec:motivation}

A deep neural network model is composed of various operators and different kinds of operators have different performance characteristics due to different ways in which hardware resources are used.
Considering whether to utilize the training process to learn the weights, the DNN operators can be roughly divided into two categories: parametric operators and non-parametric operators.
On the one hand, for parametric operators, the convolution operator (abbreviated as COP) is arguably the most important component in contemporary intelligent applications, which is utilized to perform feature extraction.
On the other hand, the non-parametric operators that are mainly based on scalar calculation (abbreviated as SOP), such as activation and element-wise addition, also provide indispensable functions.
Neural network pruning is a key optimization technique for edge intelligence, which accelerates the inference by decreasing the computational cost of operators.
However, existing pruning methods mostly focus only on optimizing COPs rather than SOPs, because the COPs are relatively more time-consuming and take up most of the execution time.
The performance of SOPs is unattended, even though the non-parametric operators are non-negligible for the DNN inference on intelligent edge devices in practice.

Let $S$ ($\bar{S}$) be the set of pruned (non-pruned) operators in a DNN model. The speedup achieved by a pruned model is:
\begin{equation}\label{equ:sppedup}
\centering
Speedup=\frac{1}{\left(1-p \left(S \right)\right)+\frac{p\left(S\right)}{a\left(S\right)}}
\end{equation}

where $p(S)$ is the percentage of the total execution time
spent on executing the operators in $S$, which implies that
$p(\bar{S}) = 1- p(S)$,
and $a(S)$ is the speedup achieved for accelerating  $S$.
Consequently, the best speedup is limited by Amdahl's law:
\begin{equation}
\centering
Speedup \leqslant \frac{1}{ 1 - p \left( S \right)
} = \frac{1}{p(\bar{S})}
\end{equation}
This shows that the speedup is always limited by $p(\bar{S}) = 1- p(S)$, which cannot benefit from the pruning optimization.

\begin{figure}[t!]
	\centering
	\includegraphics[width=0.80\linewidth]{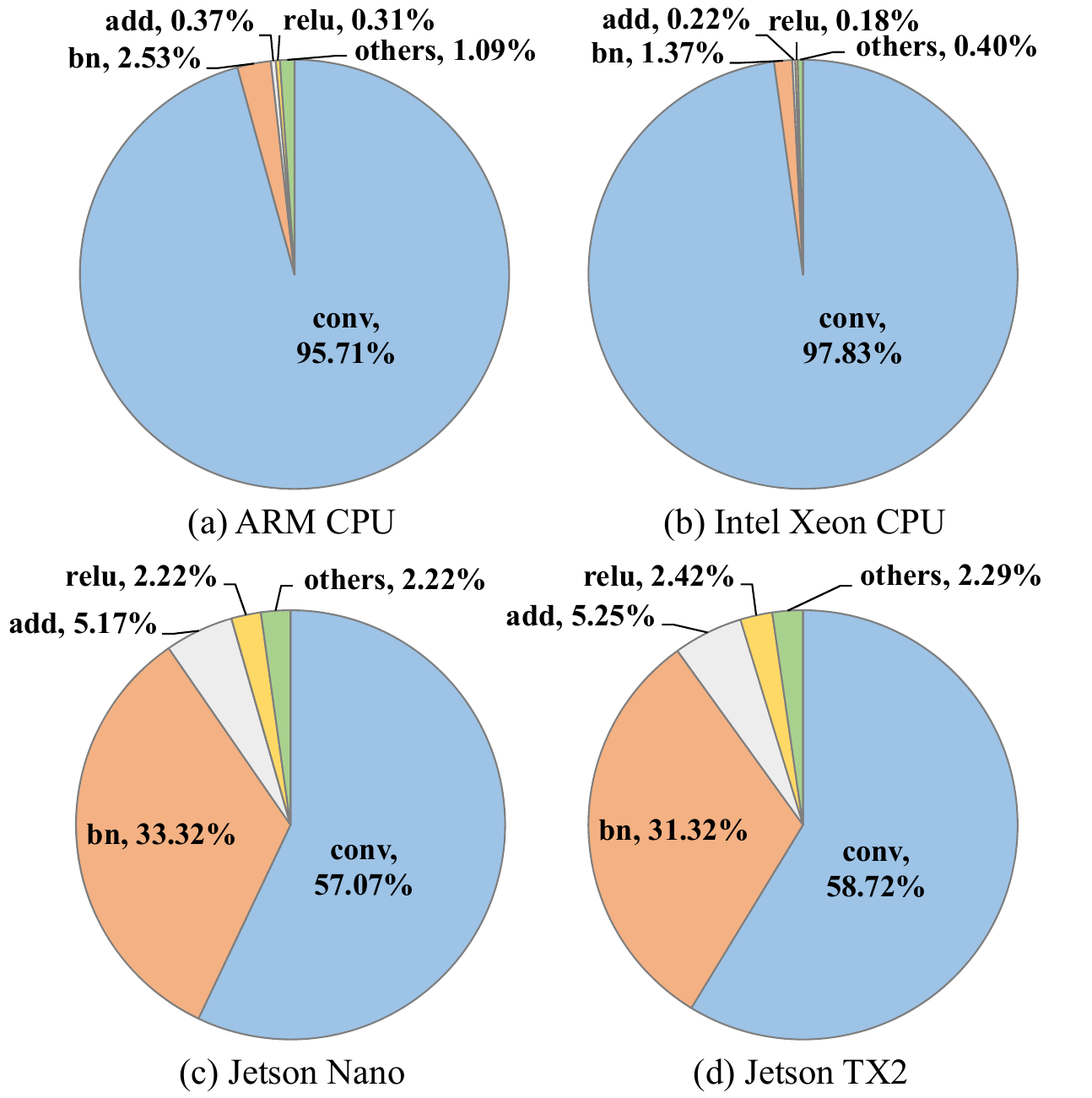}
	\caption{The performance of the ResNet-18 model on different devices. The model mainly consists of convolution (conv), batch normalization (bn), element-wise addition (add), and ReLU activation (relu) operators.}
	\label{fig:time_proportion}
\end{figure}

While preliminary studies have characterized the end-to-end performance for neural network models on intelligent edge devices~\cite{hadidi2019characterizing}, their performance characteristics (e.g., the ratio of its execution time over the total) remain unclear.
As an example, we compare the operator-level performance between general-purposed devices (Intel Xeon CPU and ARM Cortex CPU) and specialized devices (Jetson TX2, Jetson Nano) by executing the ResNet-18~\cite{he2016deep} on the ImageNet dataset~\cite{deng2009imagenet}.
Figure~\ref{fig:time_proportion} depicts the performance of ResNet-18 model on different devices.
As can be seen, for CPU platforms, concentrating on pruning COPs is a reasonable choice to accelerate DNN models because the $p(\bar{S})$ is very small.
However, for intelligent edge platforms, optimizing SOPs can also achieve considerable acceleration due as $p(\bar{S})$ becomes non-negligible, which leads to new optimization opportunities.
Although the deep learning frameworks usually optimize SOPs at the system level, there are few efforts on optimizing non-parametric operators by the optimization of model compression (e.g., pruning) in the development stage.
In this paper, we address the following challenging problem: \textit{Can we optimize non-parametric operators of neural networks in the model compression approach in a similar manner as for parametric operators?}

\begin{figure*}[ht!]
	\centering
	\includegraphics[width=0.98\linewidth]{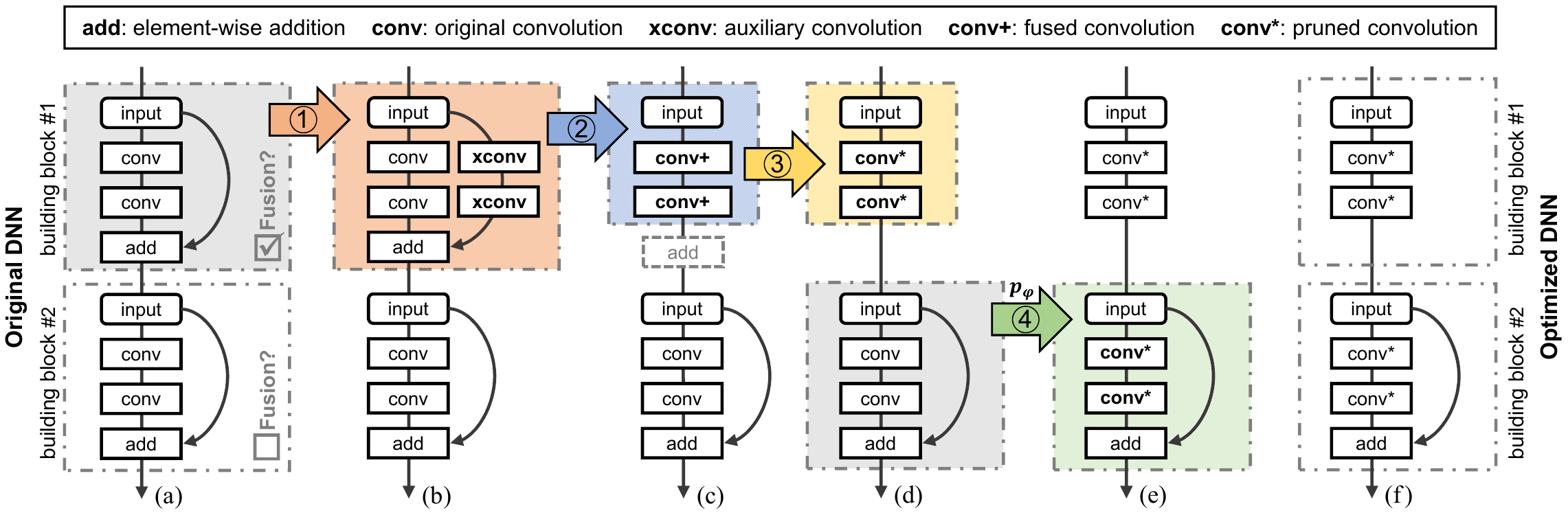}
	\caption{Overview of \FuPruner. Each arrow represents an optimization step and the dotted block in the same color indicates the corresponding optimized result. (a) to (d) describe the step-by-step optimization sequence that fuses and prunes the ``building block~\#1'', (d) to (e) describe the pruning optimization for non-fused ``building block~\#2'', and (f) shows the final optimized DNN model.}
	\vspace{-2ex}
	\label{fig:overview}
\end{figure*}

\section{Proposed Approach}
\label{sec:approach}

In this section, we will give a comprehensive introduction of \textsc{FuPruner},
and highlight the key novelty of our approach: pruning neural networks catalyzed by operator fusion.

\subsection{Overview}

Figure \ref{fig:overview} gives the workflow of our  approach and shows that how we can optimize non-parametric operators using the pruning process.
Given a pre-trained deep neural network model, \FuPruner will optimize it in terms of user-defined parameters, fusion options and pruning rates.
The fusion option determines the building blocks to be fused (marked with a ``\CheckedBox''), and the pruning rate ($p_\varphi$) determines the number of filters to be deleted for operators.
These parameters control the degree of optimization being applied, enabling the accuracy and performance tradeoffs to be made.
With reference to \ding{172}-\ding{175} in the figure, we summarize our optimization steps as follows:
\begin{enumerate}[leftmargin=2.5ex]
	\item[\ding{172}] \textbf{Inserting and Transforming Operators.}
	The non-parametric operators in the original model are difficult to be fused, which is limited by their different computation modes from the parametric operators.
	To extend the opportunities of optimization for operator fusion, we insert or transform some operators in the original model, while maintaining the equivalence before and after the model transformation.
	
	\item[\ding{173}] \textbf{Fusing Operators.}
	Unlike existing operator fusion methods that focus on decreasing the model's computational cost, we utilize operator fusion techniques to merge non-parametric operators into parametric operators so that the non-parametric operators can be pruned in later steps, which, however, may degrade the performance due to the operators inserted or transformed in step \ding{172}. Note that the architecture of the fused model is different from the original one, whereas they are equivalent.
	
	\item[\ding{174}] \textbf{Pruning Fused Operators.}
	The auxiliary operators and non-parametric operators are fused into other operators in the fused model, leading to the dilated weights of the corresponding fused operators, i.e., the fused operators have more filters than before.
    Therefore, we perform a filter pruning process  to decrease the number of such filters.

	\item[\ding{175}] \textbf{Pruning Whole Model.}
	The other operators except for the fused operators will be pruned according to the rate $p_\varphi$, so that a much smaller optimized model is obtained. We prune not only parametric operators but also non-parametric operators in the final optimized model by combining filter pruning with operator fusion techniques.

\end{enumerate}

\subsection{Aggressive Operator Fusion}
\label{sec:op_fusion}

Existing operator fusion techniques usually perform a diminishing fusion approach, which gradually reduces the number of neural network operators by fusing and transforming the execution order of operators if necessary.
On the one hand, these fusion methods are performed only in the deployment stage, whereas integrating the fusion approach into the model compression in the development stage may facilitate the model optimization.
On the other hand, the conservative fusion scheme may miss the opportunities for further optimizing models.
In this paper, we have designed an aggressive fusion scheme, which relies on inserting auxiliary operators and transforming existing operators while keeping the equivalency, to explore more optimization spaces, enabling the possibility of pruning non-parametric operators.

\subsubsection{Inserting and Transforming Convolution Operators}

\begin{figure}[t!]
	\centering
	\includegraphics[width=1\linewidth]{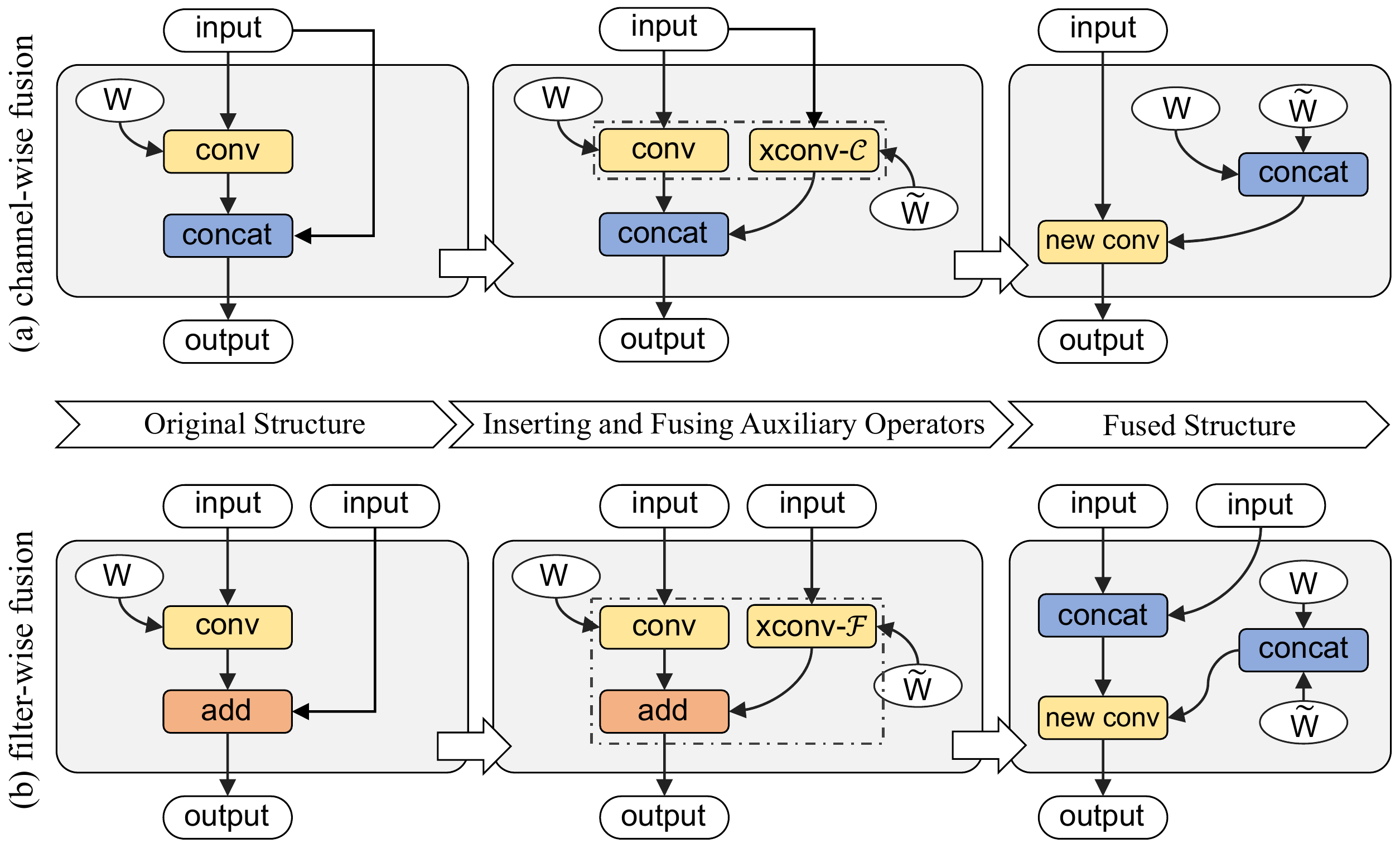}
	\caption{Inserting auxiliary convolution operators: (a)   the inserting and fusing procedure of $xconv$-$\mathcal{C}$, and (b) the procedure of $xconv$-$\mathcal{F}$.}
	\label{fig:xconv}
\end{figure}

We define two auxiliary convolution operators: an auxiliary convolution operator for channel-wise fusion (denoted as $xconv$-$\mathcal{C}$) and an auxiliary convolution operator for filter-wise fusion (denoted as $xconv$-$\mathcal{F}$).
$xconv$-$\mathcal{C}$, which is designed for the concatenation structures, merges the output of a convolution with another data, and $xconv$-$\mathcal{F}$, which is responsible for the element-wise addition structures, performs addition of the output of a convolution and another data, as shown in Figure \ref{fig:xconv}.
 $new$ $conv$ denotes a fused convolution operator, which is generated by the operators in dotted boxes.
We use the concatenation structure as an example to demonstrate the approach for channel-wise fusion.
The input of the structure is denoted by $x$ and the output by $y$.
To achieve the equivalence, the weights of the auxiliary convolution operators are composed by specific identity matrices (denoted as $\tilde{W}$), so that $x=xconv(x, \tilde{W})$.
First of all, we can represent the original structure as $y=concat(conv(x, W), x)$.
Then, we insert a $xconv$-$\mathcal{C}$ into the structure: $y = concat(conv(x, W), xconv(x, \tilde{W}))$.
By their definitions, we can fuse $conv$ and $xconv$-$\mathcal{C}$  into a $new$ $conv$ along the dimension of $C$, and consequently, obtain the fused structure: $y = new\,conv(x, concat(W, \tilde{W}))$.
As such, despite the introduced operators, the output of the neural network remains unchanged.
The approach for fusing addition structures, by fusing $add$ into $new$ $conv$ in element-wise addition structures, is similar.
The major difference between filter-wise and channel-wise fusion lies in the dimension of the concatenation operation for fusing the convolution operators.
In addition, the input of $new$ $conv$ in the fused element-wise addition structure is a concatenation of the two original $input$'s.
In practice, we can optimize common structures in deep neural networks by leveraging $xconv$-$\mathcal{C}$ and $xconv$-$\mathcal{F}$ cooperatively, as discussed below.

\begin{figure}[t!]
	\centering
	\includegraphics[width=0.9\linewidth]{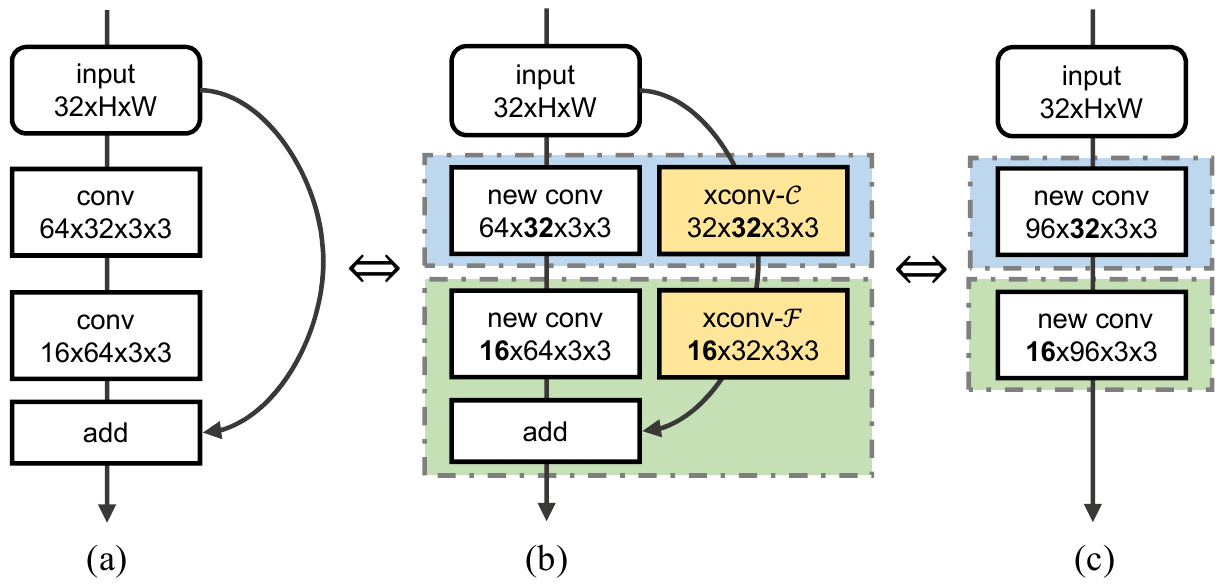}
	\caption{An example of aggressive fusion for a basic residual block: (a) original structure; (b) inserting auxiliary convolution operators (highlighted in yellow); (c) fusing original operators and auxiliary operators to new operators.}
	\label{fig:basic_res}
\end{figure}

Figure \ref{fig:basic_res} depicts the approach for fusing a basic residual block, which is a popular structure in state-of-the-art neural networks, by inserting auxiliary convolution operators.
We use $k \times c \times r \times s$ to represent the configuration of a convolution operator, where $k$ denotes the number of filters, $c$ denotes the channels, $r$ denotes the height of filters, and $s$ denotes the width of filters.
There are two 3$\times$3 convolution operators and a shortcut from input to the element-wise addition operator in the original structure.
In order to be fusible, the filter size of auxiliary operators must be the same as that for the existing convolution operators, i.e. 3$\times$3.
Moreover, $C$ in $xconv$-$\mathcal{C}$ is the same as that for the first convolution operator and  $K$ in $xconv$-$\mathcal{F}$ is  the same as  the second convolution operator.
Then, the convolution and auxiliary operators can be fused along the dimension of $C$ or $K$, as shown in Figure \ref{fig:basic_res}(c).
Thus, the element-wise addition operator in the structure is reduced.

By inserting auxiliary convolution operators, the shortcut connection can be reduced.
However, sometimes there are operators on the connection, hindering their applicability.
As such, we propose padded convolution operators, $pconv$-$\mathcal{C}$ and $pconv$-$\mathcal{F}$ for  channel-wise and filter-wise fusion.
The padded convolution operator is transformed from an existing convolution operator in neural networks by padding weights.
For example, a 1$\times$1 convolution operator can be transformed to a 3$\times$3 convolution operator by padding weights with zeros and remains to produce the same output.

\begin{figure}[t!]
	\centering
	\includegraphics[width=0.9\linewidth]{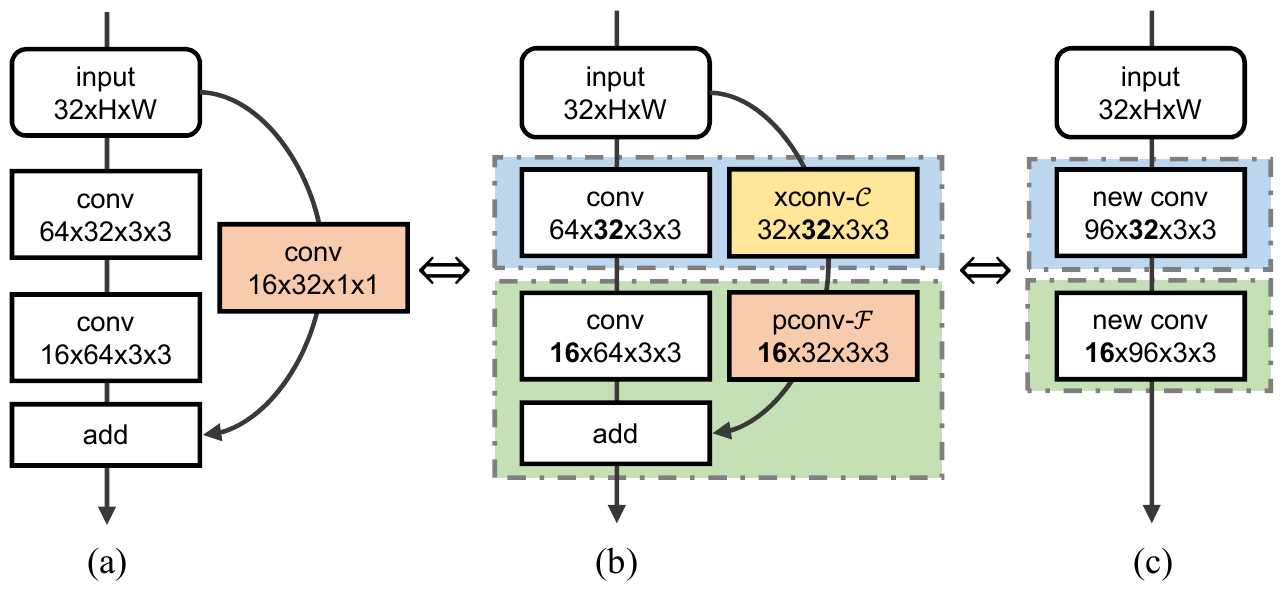}
	\caption{An example of aggressive fusion for a residual block with a projection shortcut: (a) original structure; (b) inserting auxiliary convolution (highlighted in yellow) and padding convolution operators  (highlighted in orange); (c) fusing original operators and auxiliary operators to new operators.}
	\label{fig:proj_res}
\end{figure}

Figure \ref{fig:proj_res} depicts the approach for fusing a residual block with projection shortcut by inserting auxiliary convolution operators and transforming padded convolution operators.
There is a 1$\times$1 convolution operator on the shortcut, which cannot directly be fused with other operators.
As such, we transform the 1$\times$1 convolution operator to a 3$\times$3 padded convolution operator, $pconv$-$\mathcal{F}$, which can be fused with the second original 3$\times$3 convolution operator (green area).
After that, the fusion of the structure can be completed in a similar way to the basic residual block.

\subsubsection{Adjusting Batch Normalization Operators}
Batch normalization~\cite{ioffe2015batch}, a useful technique that standardizes the inputs for each mini-batch, has been widely used in contemporary deep neural networks, which can improve the performance and stability.
Generally, the batch normalization operator (abbreviated as $bn$) can be formalized as:
\begin{equation}
\centering
BN(x)=\omega x + \lambda
\end{equation}
where
\begin{equation}
\centering
\omega=\frac{\gamma}{\sqrt{\sigma^{2}_{\mathcal{B}} +\epsilon}}
\end{equation}
\begin{equation}
\centering
\lambda = \beta - \omega\cdot\mu_{\mathcal{B}}
\end{equation}
The $\mu_{\mathcal{B}}$ and $\sigma^{2}_{\mathcal{B}}$ are the mean and variance for a mini-batch, and the $\epsilon$ is a constant for numerical stability.
The $\gamma$ and $\beta$ are the learned parameters for scale and shift.
The batch normalization can be seen as an affine transformation and the detailed algorithm can be seen in \cite{ioffe2015batch}.
The architecture of neural network that contains $bn$ operators is more complex to achieve an equivalence transformation.

\begin{figure}[t!]
	\centering
	\includegraphics[width=1\linewidth]{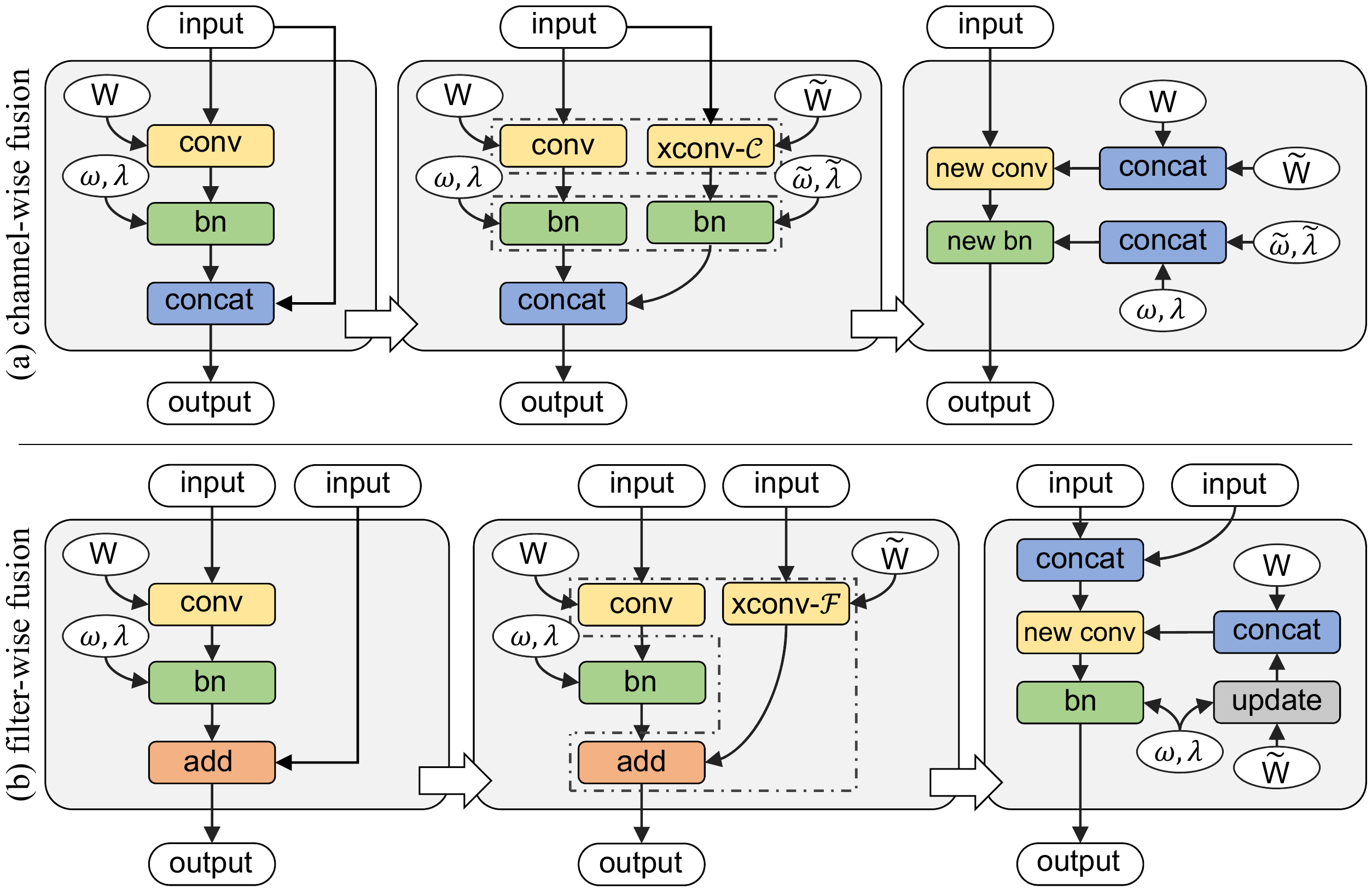}
	\caption{Disposing batch normalization operators. (a) shows the adjusting procedure of batch normalization operators with $xconv$-$\mathcal{C}$, and (b) shows the procedure with $xconv$-$\mathcal{F}$.}
	\label{fig:bn}
\end{figure}

Figure~\ref{fig:bn} depicts the procedure of disposing batch normalization operators.
For the channel-wise fusion, an auxiliary $bn$ operator ($\tilde{\omega}=1$ and $\tilde{\lambda}=0$) is inserted, which will not influence the result.
Then, the original $bn$ operator and auxiliary $bn$ operator can be fused into a $new$ $bn$ operator.
For the filter-wise fusion, the $bn$ operator remains unchanged in the fused block to achieve equivalency.
However, the $bn$ operator will disturb the result of $xconv$-$\mathcal{F}$.
As such, we update the weights of $xconv$-$\mathcal{F}$ by using the parameters of $bn$ (i.e., perform a inverse operation of $bn$ on the $\tilde{W}$).
In this way, the aggressive fusion method can be applied to the structures which have $bn$ operators.

\subsection{Dynamic Filter Pruning}
\label{sec:filter-pruning}

The fusion process is an equivalence transformation
(i,e,. the models before and after fusion are equivalent),
enabling the opportunity of pruning SOPs in the original model by fusing some SOPs into COPs.
The fused model reduces the SOPs of the model but the COPs are dilated, which is because the weights of auxiliary operators are merged into original operators.
Fortunately, we note that the involved weights in auxiliary convolutions and padded convolutions are dramatically sparse, which have more potential for compression.
As such, the fused model will be pruned to reduce the redundant weights.
In \textsc{FuPruner}, we use a soft filter pruning scheme~\cite{he2018soft}, which prunes the filters dynamically.
Unlike the classical static pruning scheme, the pruned filters in dynamic pruning scheme can be updated and recovered in the training stage, which has more model capacity and higher accuracy.

We divide the pruning process into two situations:
1) we prune the operators that involve new weights by the aggressive fusion, which prunes the model to recover the original size, i.e., the weights involved by fusion are pruned after basic pruning and the SOPs are reduced;
2) we use a pruning rate $p_\varphi$ to prune the weights of non-fused COPs.
As such, the COPs and SOPs of the neural network model are optimized simultaneously.
For each training epoch, we dynamically select the filters to be pruned according to their $\ell_2$-norm:
\begin{equation}
\centering
\left \| W_{k} \right \|_{2} = \sqrt{\sum^{c}_{t=1}\sum^{r}_{i=1}\sum^{s}_{j=1}{w^{2}_{t,i,j}}}
\end{equation}
where $W_{k}$ denotes the $k$-th filter in a convolution operator, and $w_{t,i,j}$ denotes an element of $W_{k}$ that resides in the $i$-th row and $j$-th column in the $t$-th input channel.

\begin{algorithm}[t!]
	\SetCommentSty{itshape}
	\caption{Dynamic Filter Pruning}
	\label{algo:pruning}
	\KwIn{$X$ (training data), $p_{\varphi}$ (pruning rate),
		\\ $e_{max}$ (training epoch), $W$ (original weights)}
	\KwOut{$W^{*}$ (pruned weights of model)}
	\For{$i\leftarrow 1$ \KwTo $e_{max}$\label{l-1}}{
		Update the parameters $W$ by using data $X$\label{l-2}\;
		\ForEach{$op \in model$\label{l-3}}{
			\If{$op.type = COP$\label{l-4}}{
				$n \gets $ the filter number of $op$\label{l-5}\;
				Calculate the $\ell_2$-norm for the filters\label{l-6}\;
				\eIf{$op$ is fused\label{l-7}}{
					$m \gets $ the orginal filter number of $op$\label{l-8}\;
					Zeroize the lowest $n - m$ filters\label{l-9}\;
				}{
					Zeroize the lowest $p_{\varphi} \times n$ filters\label{l-11}\;
				}
			}
		}	
	}
	Obtain the pruned model parameters $W^{*}$ from $W$\label{l-12}\;
	\Return $W^{*}$\label{l-13}\;
\end{algorithm}

The dynamic filter pruning approach is summarized in Algorithm \ref{algo:pruning}.
For each epoch, the weight parameters $W$ are updated by using training data (Line \ref{l-2}).
Then, for each COP, we obtain its filter amount $n$ and calculate the $\ell_2$-norm for all filters (Lines \ref{l-5}-\ref{l-6}).
The filters with low $\ell_2$-norm will be pruned (masked by the zero matrices).
The fused operators are pruned out of $n-m$ filters (Lines \ref{l-8}-\ref{l-9}), where $m$ indicates the filter amount of the original operators not fused, while the non-fused operators are pruned out of $p_{\varphi} \times n$ filters (Line \ref{l-11}).
For dynamic filter pruning, the weights of pruned filters  can
still be updated in the next epoch, which retains the learning capability.
Finally, the pruned model parameters $W^{*}$ are obtained from the updated $W$ according to the mask information in the last training epoch, i.e, the masked filters are removed (Lines \ref{l-12}-\ref{l-13}).

\begin{table*}[t!]
	\centering
	\caption{The specifications for the intelligent edge devices used.}
	\renewcommand\arraystretch{1.15}
	\begin{tabular}{|p{2.2cm}<{\centering}||p{2.2cm}<{\centering}|p{2.2cm}<{\centering}||p{2.2cm}<{\centering}|c|c|}
		\hline
		\textbf{Category} & \multicolumn{2}{c||}{\textbf{GPU-Based Edge Devices}} & \multicolumn{3}{c|}{\textbf{ASIC-Based Edge Devices}} \\ \hline \hline
		Platform & Jetson TX2 \cite{tx2} & Jetson Nano \cite{nano} & \begin{tabular}[c]{@{}c@{}}Coral Dev Board\\ (Edge TPU) \cite{tpu} \end{tabular} & \begin{tabular}[c]{@{}c@{}}Movidius\\ Neural Compute Stick \cite{ncs} \end{tabular} & \begin{tabular}[c]{@{}c@{}}Neural\\ Compute Stick 2 \cite{ncs2} \end{tabular} \\ \hline
		CPU & \begin{tabular}[c]{@{}c@{}}2 Denver \\ \& Cortex-A57\end{tabular} & Cortex-A57 & Cortex-A53 & N/A & N/A \\ \hline
		{\begin{tabular}[c]{@{}c@{}}ML Acceleration\\ Module\end{tabular}} & \begin{tabular}[c]{@{}c@{}}256-Core\\ Pascal GPU\end{tabular} & \begin{tabular}[c]{@{}c@{}}128-Core\\ Maxwell GPU\end{tabular} & Edge TPU & Myriad X VPU & Myriad 2 VPU \\ \hline
		{Memory} & 8 GB & 4 GB & 1 GB & N/A & N/A \\ \hline
		{\begin{tabular}[c]{@{}c@{}}Inference\\ Framework\end{tabular}} & Caffe & Caffe & TensorFlow Lite & OpenVINO & OpenVINO \\ \hline
	\end{tabular}
	\label{tab:platforms}
\vspace{-1ex}
\end{table*}

\section{Experimental Setup}
\smallskip
\noindent\textbf{Evaluation Platforms.}
We perform the procedure of fusion and pruning on a cloud server with
Intel Xeon CPUs and an NVIDIA Tesla V100 GPU.
We evaluate the performance of our approach on five representative intelligent edge devices, which contain different machine learning (ML) acceleration modules, including Jetson TX2, Jetson Nano, Coral Dev Board (Edge TPU), Movidius Neural Compute Stick (NCS), and Neural Compute Stick 2 (NCS2).
Table~\ref{tab:platforms} describes the  specifications of
these intelligent edge devices.
The optimized model will be executed on the inference framework supported by the target platform.
Especially, the Jetson TX2 and Jetson Nano run Linux-based operating systems.  We leverage the Caffe framework~\cite{jia2014caffe}, a plain deep learning framework,
to evaluate the inference performance of models on both,   so that can avoid interference of the optimizations other than ours.

\smallskip
\noindent\textbf{Benchmark Datasets.}
We evaluate our approach on two representative image classification benchmarks: CIFAR-10~\cite{krizhevsky2009learning} and ImageNet (ILSVRC2012)~\cite{deng2009imagenet}.
CIFAR-10  contains 60000 images (50000 for training and 10000 for testing), classified in ten categories.
ImageNet is a large-scale dataset, which has more than one million images classified in 1000 categories.

\smallskip
\noindent\textbf{Neural Network Models.}
In this paper, we focus on optimizing the state-of-the-art residual neural network (ResNet) models~\cite{he2016deep}.
As revealed in previous studies~\cite{he2017channel}, the ResNet models are less redundant and more challenging to compress.
Moreover, the shallower versions of ResNet models are selected, which have smaller sizes and less computation, due to the limited resources of edge devices.
Table \ref{tab:res_model} describes the architectures of the models, including ResNet-20 and ResNet-32 for CIFAR-10, and ResNet-18 and ResNet-34 for ImageNet.
In this table, the building blocks are shown in brackets and the succeeding ``$\times number$'' denotes the numbers of blocks stacked.
A convolution operator (conv) is denoted as a tuple of the size and number of its filters,
while a fully-connected operator (fc) is denoted as a number of its output dimensions.
Especially, each building block in residual neural networks except conv1 and fc ends with an element-wise addition operator (add), as shown in Figure~\ref{fig:basic_res} and Figure~\ref{fig:proj_res}.

\begin{table}[t!]
	\centering
	\caption{The architectures of the residual neural networks.}
	\renewcommand\arraystretch{1.2}
	\begin{tabular}{|c||c|c|c||c|c|c|}
		\hline
		\textbf{Layer} & \textbf{Structure} & \textbf{20} & \textbf{32} & \textbf{Structure} & \textbf{18} & \textbf{34}\\ \hline
		conv1 &
		\makecell*[c]{$\begin{bmatrix}\begin{smallmatrix}\textup{3}\times\textup{3, 16 }\end{smallmatrix}\end{bmatrix}$} &
		\makecell*[c]{$\times$1} & \makecell*[c]{$\times$1} &
		\makecell*[c]{$\begin{bmatrix}\begin{smallmatrix}\textup{7}\times\textup{7, 64 } \end{smallmatrix}\end{bmatrix}$}
		& \makecell*[c]{$\times$1} & \makecell*[c]{$\times$1} \\ \hline
		stage1 &
		\makecell*[c]{\textup{$\begin{bmatrix}\begin{smallmatrix}\textup{3}\times\textup{3, 16 }\\ \textup{3}\times\textup{3, 16 } \\  \textup{add} \end{smallmatrix}\end{bmatrix}$}} &
		\makecell*[c]{$\times$3} & \makecell*[c]{$\times$5} &
		\makecell*[c]{$\begin{bmatrix}\begin{smallmatrix}\textup{3}\times\textup{3, 64 }\\ \textup{3}\times\textup{3, 64 } \\ \textup{add} \end{smallmatrix}\end{bmatrix}$}
		& \makecell*[c]{$\times$2} & \makecell*[c]{$\times$3} \\ \hline
		stage2 &
		\makecell*[c]{$\begin{bmatrix}\begin{smallmatrix}\textup{3}\times\textup{3, 32 }\\ \textup{3}\times\textup{3, 32 } \\ \textup{add} \end{smallmatrix}\end{bmatrix}$} &
		\makecell*[c]{$\times$3} & \makecell*[c]{$\times$5} &
		\makecell*[c]{$\begin{bmatrix}\begin{smallmatrix}\textup{3}\times\textup{3, 128 }\\ \textup{3}\times\textup{3, 128 } \\ \textup{add} \end{smallmatrix}\end{bmatrix}$}
		& \makecell*[c]{$\times$2} & \makecell*[c]{$\times$4}\\ \hline
		stage3 &
		\makecell*[c]{$\begin{bmatrix}\begin{smallmatrix}\textup{3}\times\textup{3, 64 }\\ \textup{3}\times\textup{3, 64 } \\ \textup{add} \end{smallmatrix}\end{bmatrix}$} &
		\makecell*[c]{$\times$3} & \makecell*[c]{$\times$5} &
		\makecell*[c]{$\begin{bmatrix}\begin{smallmatrix}\textup{3}\times\textup{3, 256 }\\ \textup{3}\times\textup{3, 256 } \\ \textup{add} \end{smallmatrix}\end{bmatrix}$}
		& \makecell*[c]{$\times$2} & \makecell*[c]{$\times$6} \\ \hline
		stage4 & - & - & - &
		\makecell*[c]{$\begin{bmatrix}\begin{smallmatrix}\textup{3}\times\textup{3, 512 }\\ \textup{3}\times\textup{3, 512 } \\ \textup{add} \end{smallmatrix}\end{bmatrix}$}
		& \makecell*[c]{$\times$2} & \makecell*[c]{$\times$3} \\ \hline
		fc &
		\makecell*[c]{$\begin{bmatrix}\begin{smallmatrix}\textup{10 } \end{smallmatrix}\end{bmatrix}$} &
		\makecell*[c]{$\times$1} & \makecell*[c]{$\times$1} &
		\makecell*[c]{$\begin{bmatrix}\begin{smallmatrix}\textup{1000 } \end{smallmatrix}\end{bmatrix}$}
		& \makecell*[c]{$\times$1} & \makecell*[c]{$\times$1} \\ \hline
	\end{tabular}
	\label{tab:res_model}
\end{table}

\smallskip
\noindent\textbf{Optimization Setting.}
The fusion and pruning of \textsc{Fu\-Pruner} approach are based on the PyTorch framework~\cite{paszke2019pytorch}.
In the pruning stage, its default data argumentation strategy is utilized.
Meanwhile, we use the same setting of hype-parameters and training schedules as \cite{he2018soft}.
The fusion options are denoted as ``$x/n$'', where $n$ is the number of stages contained in the model, and $x$ is the number of stages to be fused.
For simplicity, we set the stages to be fused in order according to the stage numbers.
We prune the fused convolution operators of neural networks, which retains exactly the same number of filters as before fusion.
We set the same continued pruning rates, from 0 to 0.3, for all non-fused convolution operators.
As such, the fusion option and pruning rate are utilized to balance accuracy and performance.

\begin{table*}[!th]
	\centering
	\caption{The accuracy of pruned residual neural network models on the CIFAR-10 dataset.}
	\renewcommand\arraystretch{1.12}
	\begin{tabular}{|c|p{2.5cm}<{\centering}|p{2.5cm}<{\centering}|p{2.5cm}<{\centering}|p{1.8cm}<{\centering}|c|c|}
		\hline
		\textbf{Model} & \textbf{Method} & \textbf{\begin{tabular}[c]{@{}c@{}}Top-1 Acc.\\ Baseline (\%)\end{tabular}} & \textbf{\begin{tabular}[c]{@{}c@{}}Top-1 Acc.\\ Pruned (\%)\end{tabular}} & \textbf{\begin{tabular}[c]{@{}c@{}}Top-1 Acc.\\ Drop (\%)\end{tabular}} & \textbf{\begin{tabular}[c]{@{}c@{}} Fusion Option \end{tabular}} & \textbf{\begin{tabular}[c]{@{}c@{}} Pruning Rate (\%)\end{tabular}} \\ \hline
		\multirow{11}{*}{ResNet-20} & MIL \cite{dong2017more} & 91.53 & 91.43 & 0.10 & N/A & N/A \\
		& SFP-0.1 \cite{he2018soft} & 92.20 & 92.24 & -0.04 & N/A & 10.0\% \\
		& SFP-0.2 \cite{he2018soft} & 92.20 & 91.20 & 1.00 & N/A & 20.0\% \\
		& SFP-0.3 \cite{he2018soft} & 92.20 & 90.83 & 1.37 & N/A & 30.0\% \\ \cdashline{2-7}[2pt/2pt]
		
		& \FuPruner-1/3 & 92.60 ($\pm$0.27) & \textbf{92.86 ($\pm$0.02)} & \textbf{-0.26} & 1/3 & 0.0\% \\
		& \FuPruner-2/3 & 92.60 ($\pm$0.27) & \textbf{92.58 ($\pm$0.10)} & \textbf{0.02} & 2/3 & 0.0\% \\
		& \FuPruner-3/3 & 92.60 ($\pm$0.27) & \textbf{92.34 ($\pm$0.05)} & \textbf{0.26} & 3/3 & 0.0\% \\ \cdashline{2-7}[2pt/2pt]
		& \FuPruner-1/3-0.1 & 92.60 ($\pm$0.27) & \textbf{92.77 ($\pm$0.19)} & \textbf{-0.17} & 1/3 & 10.0\% \\
		& \FuPruner-1/3-0.2 & 92.60 ($\pm$0.27) & \textbf{92.28 ($\pm$0.13)} & \textbf{0.32} & 1/3 & 20.0\% \\
		& \FuPruner-1/3-0.3 & 92.60 ($\pm$0.27) & \textbf{91.65 ($\pm$0.16)} & \textbf{0.95} & 1/3 & 30.0\% \\
		& \FuPruner-3/3-0.3 & 92.60 ($\pm$0.27) & \textbf{91.40 ($\pm$0.05)} & \textbf{1.20} & 3/3 & 30.0\% \\ \hline
		\multirow{11}{*}{ResNet-32} & MIL \cite{dong2017more} & 92.33 & 90.74 & 1.59 & N/A & N/A \\
		& SFP-0.1 \cite{he2018soft} & 92.63 & 93.22 & -0.59 & N/A & 10.0\% \\
		& SFP-0.2 \cite{he2018soft} & 92.63 & 90.63 & 2.00 & N/A & 20.0\% \\
		& SFP-0.3 \cite{he2018soft} & 92.63 & 90.08 & 2.55 & N/A & 30.0\% \\ \cdashline{2-7}[2pt/2pt]
		& \FuPruner-1/3 & 93.41 ($\pm$0.16) & \textbf{93.29 ($\pm$0.09)} & \textbf{0.12} & 1/3 & 0.0\% \\
		& \FuPruner-2/3 & 93.41 ($\pm$0.16) & \textbf{92.40 ($\pm$0.09)} & \textbf{1.01} & 2/3 & 0.0\% \\
		& \FuPruner-3/3 & 93.41 ($\pm$0.16) & \textbf{92.32 ($\pm$0.08)} & \textbf{1.09} & 3/3 & 0.0\% \\ \cdashline{2-7}[2pt/2pt]
		& \FuPruner-1/3-0.1 & 93.41 ($\pm$0.16) & \textbf{93.33 ($\pm$0.02)} & \textbf{0.08} & 1/3 & 10.0\% \\
		& \FuPruner-1/3-0.2 & 93.41 ($\pm$0.16) & \textbf{93.23 ($\pm$0.14)} & \textbf{0.18} & 1/3 & 20.0\% \\
		& \FuPruner-1/3-0.3 & 93.41 ($\pm$0.16) & \textbf{92.55 ($\pm$0.26)} & \textbf{0.86} & 1/3 & 30.0\% \\
		& \FuPruner-3/3-0.3 & 93.41 ($\pm$0.16) & \textbf{91.94 ($\pm$0.07)} & \textbf{1.47} & 3/3 & 30.0\% \\ \hline
	\end{tabular}
	\label{tab:result-cifar}
\end{table*}

\begin{table*}[ht!]
	\renewcommand\arraystretch{1.12}
	\centering
	\caption{The accuracy of pruned residual neural network models on the ImageNet dataset.}
	\begin{tabular}{|c|p{2.6cm}<{\centering}|c|p{2.25cm}<{\centering}|c|c|p{2.25cm}<{\centering}|c|}
		\hline
		\textbf{Model} & \textbf{Method} & \textbf{\begin{tabular}[c]{@{}c@{}}Top-1 Acc.\\ Baseline (\%)\end{tabular}} & \textbf{\begin{tabular}[c]{@{}c@{}}Top-1 Acc.\\ Pruned (\%)\end{tabular}} & \textbf{\begin{tabular}[c]{@{}c@{}}Top-1 Acc.\\ Drop (\%)\end{tabular}} & \textbf{\begin{tabular}[c]{@{}c@{}}Top-5 Acc.\\ Baseline (\%)\end{tabular}} & \textbf{\begin{tabular}[c]{@{}c@{}}Top-5 Acc.\\ Pruned (\%)\end{tabular}} & \textbf{\begin{tabular}[c]{@{}c@{}}Top-5 Acc.\\ Drop (\%)\end{tabular}} \\ \hline
		\multirow{9}{*}{ResNet-18} & MIL \cite{dong2017more} & 69.98 & 66.33 & 3.65 & 89.24 & 86.94 & 2.3 \\
		& SFP-0.3 \cite{he2018soft} & 70.23 & 60.79 & 9.44 & 89.51 & 83.11 & 6.4 \\
		& ASFP-0.3 \cite{he2019asymptotic} & 70.23 & 68.02 & 2.21 & 89.51 & 88.19 & 1.32 \\ \cdashline{2-8}[2pt/2pt]
		& {XNOR-Net++~\cite{DBLP:conf/bmvc/BulatT19}} & 69.30 & 57.10 & 12.2 & 89.20 & 79.90 & 9.30 \\
		& {Bi-Real Net~\cite{liu2020bi}} & 69.30 & 56.40 & 12.9 & 89.20 & 79.50 & 9.70 \\
		& {CI-BCNN~\cite{wang2019learning}} & 69.30 & 59.90 & 9.40 & 89.20 & 84.18 & 5.02 \\		
		\cdashline{2-8}[2pt/2pt]
		& \FuPruner-1/4 & 69.76 & \textbf{70.97 ($\pm$0.03)} & \textbf{-1.21} & 89.08 & \textbf{89.85 ($\pm$0.03)} & \textbf{-0.77} \\
		& \FuPruner-2/4 & 69.76 & \textbf{70.75 ($\pm$0.07)} & \textbf{-0.99} & 89.08 & \textbf{89.81 ($\pm$0.03)} & \textbf{-0.73} \\
		& \FuPruner-3/4 & 69.76 & \textbf{70.61 ($\pm$0.01)} & \textbf{-0.85} & 89.08 & \textbf{89.70 ($\pm$0.06)} & \textbf{-0.62} \\
		& \FuPruner-4/4 & 69.76 & \textbf{70.39 ($\pm$0.44)} & \textbf{-0.63} & 89.08 & \textbf{89.55 ($\pm$0.17)} & \textbf{-0.47} \\ \cdashline{2-8}[2pt/2pt]
		& \FuPruner-4/4-0.2 & 69.76 & \textbf{69.69 ($\pm$0.25)} & \textbf{0.07} & 89.08 & \textbf{89.35 ($\pm$0.08)} & \textbf{-0.27} \\
		& \FuPruner-4/4-0.3 & 69.76 & \textbf{68.24 ($\pm$0.45)} & \textbf{1.52} & 89.08 & \textbf{88.21 ($\pm$0.27)} & \textbf{0.87} \\ \hline
		\multirow{10}{*}{ResNet-34} & MIL \cite{dong2017more} & 73.42 & 72.99 & 0.43 & 91.36 & 91.19 & 0.17 \\
		& PFEC \cite{DBLP:conf/iclr/0022KDSG17} & 73.23 & 72.17 & 1.06 & - & - & - \\
		& SFP-0.3 \cite{he2018soft} & 73.92 & 72.29 & 1.63 & 91.62 & 90.90 & 0.72 \\
		& ASFP-0.3 \cite{he2019asymptotic} & 73.92 & 72.53 & 1.39 & 91.62 & 91.04 & 0.58 \\ \cdashline{2-8}[2pt/2pt]
		& {Bi-Real Net~\cite{liu2020bi}} & 73.30 & 62.20 & 11.1 & 91.30 & 83.90 & 7.40 \\
		& {CI-BCNN~\cite{wang2019learning}} & 73.30 & 64.93 & 8.37 & 91.30 & 86.61 & 4.69 \\ \cdashline{2-8}[2pt/2pt]	
		& \FuPruner-1/4 & 73.32 & \textbf{74.28 ($\pm$0.11)} & \textbf{-0.96} & 91.42 & \textbf{91.90 ($\pm$0.07)} & \textbf{-0.48} \\
		& \FuPruner-2/4 & 73.32 & \textbf{74.23 ($\pm$0.04)} & \textbf{-0.91} & 91.42 & \textbf{91.76 ($\pm$0.04)} & \textbf{-0.34} \\
		& \FuPruner-3/4 & 73.32 & \textbf{73.85 ($\pm$0.03)} & \textbf{-0.53} & 91.42 & \textbf{91.61 ($\pm$0.04)} & \textbf{-0.19} \\
		& \FuPruner-4/4 & 73.32 & \textbf{73.10 ($\pm$0.05)} & \textbf{0.22} & 91.42 & \textbf{91.23 ($\pm$0.03)} & \textbf{0.19} \\ \cdashline{2-8}[2pt/2pt]
		& \FuPruner-4/4-0.2 & 73.32 & \textbf{72.86 ($\pm$0.07)} & \textbf{0.46} & 91.42 & \textbf{91.13 ($\pm$0.08)} & \textbf{0.29} \\
		& \FuPruner-4/4-0.3 & 73.32 & \textbf{72.14 ($\pm$0.05)} & \textbf{1.18} & 91.42 & \textbf{90.66 ($\pm$0.02)} & \textbf{0.76} \\ \hline
	\end{tabular}
	\label{tab:result-imagenet}
	\vspace{-2ex}
\end{table*}

\section{Evaluation}
\label{sec:eval}

In this section, we demonstrate the effectiveness of our proposed approach, which successfully optimizes DNN models with a negligible accuracy loss.
Specifically, we focus on answering two research questions:

\begin{itemize} [leftmargin=2.5ex]
	\item \textbf{RQ1.} Can \textsc{FuPruner} prune the DNN models effectively with a negligible accuracy loss?
	\item \textbf{RQ2.} Can \textsc{FuPruner} accelerate the end-to-end inference of deep neural networks on intelligent edge devices?
\end{itemize}

\subsection{RQ1. The Model Accuracy}
We compare \textsc{FuPruner} with state-of-the-art acceleration methods, including MIL~\cite{dong2017more}, PFEC~\cite{DBLP:conf/iclr/0022KDSG17}, SFP~\cite{he2018soft} and ASFP~\cite{he2019asymptotic}.
For fairness, the accuracy numbers are directly cited from the corresponding papers.
While the architecture of a baseline model is the same for all methods, the accuracy numbers are slightly different due to different experimental hyper-parameters, such as learning rate and data augmentation.
As such, we report not only the accuracy of optimized models but also the baseline models.
We use ``Acc. Drop'' to illustrate the accuracy dropping of the optimized model compared to the original model.
Furthermore,
we run each experiment three times and report the result as ($mean \pm std$), where $mean$ represents the average and $std$  the standard deviation,
and all the accuracies of compressed models are evaluated on the cloud server with the same environment.

We evaluate our approach on the CIFAR-10 dataset by using ResNet-20 and ResNet-32 models, as shown in Table \ref{tab:result-cifar}.
The top-1 accuracy of each method is reported in the table.
Besides classical filter pruning methods, we also compare the proposed approach with MIL~\cite{dong2017more}, an acceleration method that involves new efficient structures into the original neural network models.
We perform two pruning approaches:
1) conservative pruning: we only prune the filters of dilated convolution operators incurred by aggressive fusion, which is denoted as ``\textsc{FuPruner}-(fusion option)'', according to the fusion option;
2) continued pruning: we further prune non-fused convolution operators based on the conservative pruned models to achieve simultaneous optimization for SOPs and COPs, which is denoted as ``\textsc{FuPruner}-(fusion option)-(continued pruning rate)''.
For instance, the ``\textsc{FuPruner}-1/3'' means that we fuse building blocks in the first stage of the neural network and then prune the fused convolution operators to recover the size of the model.
The ``\textsc{FuPruner}-1/3-0.3'' means that we prune non-fused convolution operators of the neural network with the $p_\varphi=0.3$ based on the model of ``\textsc{FuPruner}-1/3''.
Firstly, \textsc{FuPruner} can prune SOPs with a negligible accuracy loss.
For example, the optimized ResNet-20 model by ``\textsc{FuPruner-1/3}'' fuses and prunes about 1/3 building blocks without any accuracy loss, and the model by ``\textsc{FuPruner-3/3}'', which fuses and prunes all the element-wise addition operators from stage1 to stage3, drops only by $0.26\%$ in accuracy.
When performing the continued pruning approach that further prunes COPs, the performance of \textsc{FuPruner} also outperforms other state-of-the-art methods.
As can be seen, our approach exhibits less accuracy loss compared to other methods for the same pruning rate.
The accuracies of some pruned models, such as the ResNet-20 models with ``SFP-0.1'' \cite{he2018soft} and ``\textsc{FuPruner-1/3}'', are even better than the baseline model. This shows that the filter pruning approach has a regularization effect and can reduce the over-fitting of neural network models.

We also evaluate our approach on ImageNet, a large-scale dataset, by using ResNet-18 and ResNet-34 models, as shown in Table \ref{tab:result-imagenet}.
We report the top-1 accuracy and top-5 accuracy for each method
and the results on ImageNet are similar to those on CIFAR-10.
Our approach can prune SOPs without any accuracy loss in most cases.
The optimized ResNet-34 model by ``\textsc{FuPruner}-4/4'', which fuses all building blocks, only drops by $0.22\%$ in top-1 accuracy and $0.19\%$ in top-5 accuracy.
Furthermore, the models optimized by continued pruning achieve better performance compared to other methods.
These experimental results show the effectiveness of \textsc{FuPruner}, which prunes SOPs and COPs simultaneously and achieves comparable performance as original models.
We have also compared our approach with several state-of-the-art neural network binarization methods, including  XNOR-Net++~\cite{DBLP:conf/bmvc/BulatT19}, Bi-Real~Net~\cite{liu2020bi}, and CI-BCNN~\cite{wang2019learning}.
While the binarized models enjoy the potential of storage compression and acceleration, their performance degradation is unsatisfactory for real intelligent applications.

\begin{figure*}[t!]
	\centering
	\includegraphics[width=0.98\linewidth]{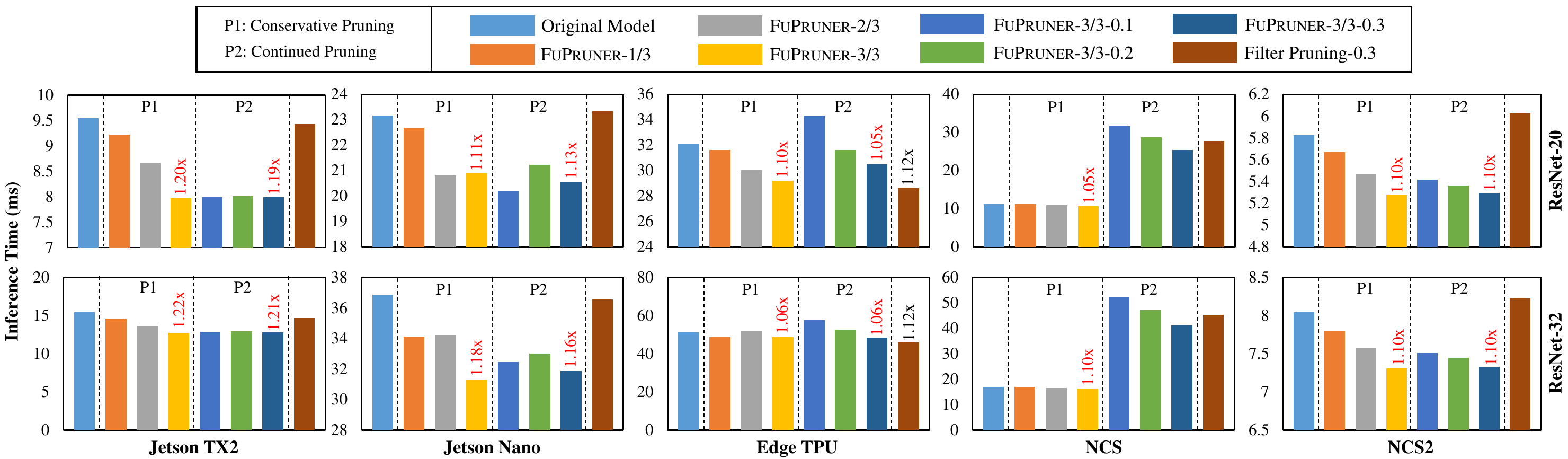}
	\caption{Comparing the inference performance results among the original model, the models optimized by \FuPruner, and the models optimized by classic filter pruning on CIFAR-10. The left-most blue bar denotes the inference time of the original DNN model, the three bars in P1 denote the conservative pruning by \FuPruner, and the three bars in P2 denote the continued pruning. The above-bar numbers denote the relative speedups over the original model.}
	\label{fig:speedup-cifar}
\end{figure*}

\begin{table*}[t!]
\caption{The best speedups of the optimized neural network models on the ImageNet dataset.}
\centering
\renewcommand\arraystretch{1.15}
\begin{tabular}{|c|c||c|c|c|c|c|c|c||c|c|c|}
\hline
\multirow{3}{*}{\textbf{Model}} & \multirow{3}{*}{\textbf{Device}} & \multicolumn{9}{c|}{\textbf{\FuPruner}} & \textbf{Ref-0.3} \\ \cline{3-12}
 &  & \multicolumn{7}{c||}{\textbf{Speedups (for Some Selected Combinations)}} & \multirow{2}{*}{\textbf{Best Param.}} & \multirow{2}{*}{\textbf{Speedup}} & \multirow{2}{*}{\textbf{Speedup}} \\ \cline{3-9}
 &  & \textbf{1/4} & \textbf{2/4} & \textbf{3/4} & \textbf{4/4} & \textbf{4/4-0.1} & \textbf{4/4-0.2} & \textbf{4/4-0.3} &  &  &  \\ \hline
\multirow{5}{*}{\rotatebox{0}{ResNet-18}} & Jetson TX2 & 1.02$\times$ & 1.06$\times$ & 1.09$\times$ & 1.12$\times$ & 1.15$\times$ & 1.22$\times$ & 1.33$\times$ & 3/4-0.3 & \textbf{1.34$\times$} & 1.30$\times$ \\ \cline{2-12}
 & Jetson Nano & 1.02$\times$ & 1.06$\times$ & 1.07$\times$ & 1.13$\times$ & 1.14$\times$ & 1.21$\times$ & 1.35$\times$ & 3/4-0.3 & \textbf{1.35$\times$} & 1.29$\times$ \\ \cline{2-12}
 & Edge TPU & 1.02$\times$ & 1.02$\times$ & 1.03$\times$ & 1.03$\times$ & 1.04$\times$ & 1.15$\times$ & 1.30$\times$ & 1/4-0.3 & \textbf{1.43$\times$} & 1.42$\times$ \\ \cline{2-12}
 & NCS & 1.00$\times$ & 1.01$\times$ & 1.02$\times$ & 1.02$\times$ & 0.18$\times$ & 0.22$\times$ & 0.23$\times$ & 4/4 & \textbf{1.02$\times$} & 0.21$\times$ \\ \cline{2-12}
 & NCS2 & 1.01$\times$ & 1.03$\times$ & 1.04$\times$ & 1.06$\times$ & 1.01$\times$ & 1.10$\times$ & 1.19$\times$ & 2/4-0.3 & \textbf{1.24$\times$} & 1.22$\times$ \\ \hline
\multirow{5}{*}{\rotatebox{0}{ResNet-34}} & Jetson TX2 & 1.02$\times$ & 1.04$\times$ & 1.07$\times$ & 1.09$\times$ & 1.12$\times$ & 1.19$\times$ & 1.29$\times$ & 2/4-0.3 & \textbf{1.33$\times$} & 1.33$\times$ \\ \cline{2-12}
 & Jetson Nano & 1.02$\times$ & 1.05$\times$ & 1.09$\times$ & 1.10$\times$ & 1.11$\times$ & 1.22$\times$ & 1.33$\times$ & 3/4-0.3 & \textbf{1.37$\times$} & 1.25$\times$ \\ \cline{2-12}
 & Edge TPU & 1.02$\times$ & 1.01$\times$ & 1.01$\times$ & 1.03$\times$ & 1.02$\times$ & 1.13$\times$ & 1.28$\times$ & 1/4-0.3 & \textbf{1.46$\times$} & 1.46$\times$ \\ \cline{2-12}
 & NCS & 1.00$\times$ & 1.01$\times$ & 1.01$\times$ & 1.01$\times$ & 0.19$\times$ & 0.23$\times$ & 0.23$\times$ & 4/4 & \textbf{1.01$\times$} & 0.21$\times$ \\ \cline{2-12}
 & NCS2 & 1.01$\times$ & 1.03$\times$ & 1.06$\times$ & 1.07$\times$ & 1.04$\times$ & 1.15$\times$ & 1.25$\times$ & 2/4-0.3 & \textbf{1.33$\times$} & 1.32$\times$ \\ \hline
\end{tabular}
\label{tab:speed-imagenet}
\end{table*}

\subsection{RQ2. The Realistic Acceleration}
We address RQ2 by evaluating the performance of single-batch inference on representative intelligent edge devices.
We compare  the original model with the models optimized by \FuPruner with different optimization settings.
The inference time reported of a model is the average of 1000 runs.

{\textit{1) Results on CIFAR-10.}}
Figure~\ref{fig:speedup-cifar} depicts the results on CIFAR-10 dataset.
As for optimized models by \FuPruner, we set fusion option~$\in\{1/3, 2/3, 3/3\}$  for conservative pruning (P1), and we set fusion option = $3/3$ and $p_{\varphi}\in \{0.1,0.2,0.3\}$ for continued pruning (P2).
Moreover, the classic filter pruning approach with the rate = $0.3$ is used as the reference.
There are three observations.
First, the fused models with conservative pruning outperforms the original models, providing the evidence for superior performance benefits obtainable from non-parametric operator pruning.
Second, the optimized models by continued pruning cannot always yield acceleration, and may even lead to performance degradation, with, e.g., the continued pruning results on NCS.
This is because the convolution operators with special sizes have more efficient implementations in deep learning frameworks.
For example, the convolution operators with 64 filters often perform faster than those with 63 filters, even though the latter have fewer filters, because the regular data size is more conducive to the optimizations such as parallelization.
Finally, \FuPruner outperforms classic filter pruning approaches, which prune only COPs, with the pruning rate = 0.3, in most cases.
The classic filter pruning approaches prune a little more convolution operators than our approach (are thus
faster  on some devices such as Edge TPU), because the fused operators will not continue to be pruned in \FuPruner.

{\textit{2) Results on ImageNet.}}
Table~\ref{tab:speed-imagenet} gives the best speedup over the original model for each optimization combination of (fusion option, $p_\varphi$) in $\{1/4,2/4,3/4,4/4\}\times\{0,0.1,0.2,0.3\}$ on ImageNet.
We report some results obtained by \FuPruner with different optimization parameters,  the best optimization parameter
values (Best Param.) and their speedups achieved.
Moreover, we use the classic filter pruning approach with the pruning rate = 0.3 as a reference in the last column (Ref-0.3).
The results demonstrate the flexibility of \FuPruner, that is, we can get the best optimized model by adjusting the fusion option and pruning rate.
Compared with the original model, the optimized models by \FuPruner achieve significant performance improvements on most evaluated edge platforms.
In addition, the accuracy of optimized model by \FuPruner is higher than the corresponding reference model (Ref-0.3), as revealed in Table~\ref{tab:result-imagenet}.
As such, the performance of our approach, which has the less accuracy loss, outperforms the classic filter pruning approaches.
One observation is that the speedups of the pruned models by continued pruning are more noticeable than those on CIFAR-10, which is attributed to the larger sizes of input images in the ImageNet dataset.
These results show that \FuPruner provides a flexible optimization framework and can achieve realistic performance improvements.

\begin{table*}[ht!]
\renewcommand\arraystretch{1.15}
\centering
\caption{{The performance results for the optimized neural network models on the CIFAR-10 dataset.}}
\begin{tabular}{|c|c|c|c||c|c|c|c|c|c||c|c|c|c|c|c|}
\hline
\multirow{2}{*}{\textbf{Model}} & \multicolumn{2}{c|}{\multirow{2}{*}{\textbf{\begin{tabular}[c]{@{}c@{}}Optimization\\ Param.\end{tabular}}}} & \multirow{2}{*}{\textbf{\begin{tabular}[c]{@{}c@{}}Top-1 \\ Acc. (\%)\end{tabular}}} & \multicolumn{6}{c||}{\textbf{Inference Times on Jetson TX2 (\textit{ms})}} & \multicolumn{6}{c|}{\textbf{Inference Times on Intel NCS2 (\textit{ms})}} \\ \cline{5-16}
 & \multicolumn{2}{c|}{} &  & \textbf{Sys.} & \textbf{Conv} & \textbf{Add} & \textbf{BN} & \textbf{Others} & \textbf{Total} & \textbf{Sys.} & \textbf{Conv} & \textbf{Add} & \textbf{BN} & \textbf{Others} & \textbf{Total} \\ \hline
\multirow{9}{*}{\rotatebox{90}{ResNet-20}} & \multicolumn{2}{c|}{Baseline} & 92.60 & 1.81 & 2.22 & 0.74 & 4.41 & 0.36 & \textbf{9.55} & 2.10 & 0.86 & 0.54 & 1.92 & 0.40 & \textbf{5.83} \\ \cline{2-16}
 & \multirow{8}{*}{\rotatebox{90}{\FuPruner}} & 1/3 & 92.86 & 1.63 & 2.17 & 0.51 & 4.48 & 0.44 & \textbf{9.22} & 2.06 & 0.86 & 0.34 & 1.93 & 0.48 & \textbf{5.67} \\ \cline{3-16}
 &  & 2/3 & 92.58 & 1.58 & 2.10 & 0.25 & 4.25 & 0.50 & \textbf{8.67} & 2.07 & 0.84 & 0.17 & 1.84 & 0.55 & \textbf{5.47} \\ \cline{3-16}
 &  & 3/3 & 92.34 & 1.44 & 1.97 & - & 3.98 & 0.57 & \textbf{7.97} & 2.05 & 0.82 & - & 1.79 & 0.63 & \textbf{5.28} \\ \cline{3-16}
 &  & 3/3-0.1 & 92.41 & 1.44 & 1.99 & - & 4.00 & 0.56 & \textbf{7.99} & 2.08 & 0.81 & - & 1.91 & 0.62 & \textbf{5.42} \\ \cline{3-16}
 &  & 3/3-0.2 & 91.92 & 1.42 & 1.98 & - & 4.03 & 0.58 & \textbf{8.01} & 2.04 & 0.81 & - & 1.89 & 0.62 & \textbf{5.36} \\ \cline{3-16}
 &  & 3/3-0.3 & 91.40 & 1.46 & 1.98 & - & 3.98 & 0.57 & \textbf{7.99} & 2.05 & 0.79 & - & 1.83 & 0.62 & \textbf{5.30} \\ \cline{3-16}
 &  & 3/3* & 92.34 & 0.60 & 1.86 & - & - & 0.57 & \textbf{3.03} & 1.14 & 0.88 & - & - & 0.16 & \textbf{2.18} \\ \cline{3-16}
 &  & 3/3-0.3* & 91.40 & 0.63 & 1.89 & - & - & 0.59 & \textbf{3.11} & 1.15 & 0.85 & - & - & 0.17 & \textbf{2.17} \\ \hline
\multirow{9}{*}{\rotatebox{90}{ResNet-32}} & \multicolumn{2}{c|}{Baseline} & 93.41 & 3.32 & 3.40 & 1.24 & 7.02 & 0.52 & \textbf{15.49} & 2.22 & 1.37 & 0.89 & 3.02 & 0.54 & \textbf{8.05} \\ \cline{2-16}
 & \multirow{8}{*}{\rotatebox{90}{\FuPruner}} & 1/3 & 93.29 & 2.73 & 3.37 & 0.83 & 7.06 & 0.64 & \textbf{14.64} & 2.19 & 1.38 & 0.55 & 3.01 & 0.68 & \textbf{7.81} \\ \cline{3-16}
 &  & 2/3 & 92.40 & 2.18 & 3.34 & 0.43 & 6.93 & 0.76 & \textbf{13.63} & 2.23 & 1.35 & 0.27 & 2.94 & 0.80 & \textbf{7.58} \\ \cline{3-16}
 &  & 3/3 & 92.32 & 1.92 & 3.23 & - & 6.71 & 0.88 & \textbf{12.74} & 2.19 & 1.33 & - & 2.87 & 0.91 & \textbf{7.31} \\ \cline{3-16}
 &  & 3/3-0.1 & 92.60 & 2.40 & 3.21 & - & 6.44 & 0.86 & \textbf{12.91} & 2.17 & 1.33 & - & 3.09 & 0.92 & \textbf{7.51} \\ \cline{3-16}
 &  & 3/3-0.2 & 92.41 & 2.46 & 3.19 & - & 6.48 & 0.85 & \textbf{12.98} & 2.21 & 1.31 & - & 3.03 & 0.90 & \textbf{7.45} \\ \cline{3-16}
 &  & 3/3-0.3 & 91.94 & 2.27 & 3.17 & - & 6.49 & 0.85 & \textbf{12.79} & 2.18 & 1.28 & - & 2.97 & 0.89 & \textbf{7.33} \\ \cline{3-16}
 &  & 3/3* & 92.32 & 0.93 & 2.95 & - & - & 0.85 & \textbf{4.73} & 1.75 & 1.42 & - & - & 0.16 & \textbf{3.33} \\ \cline{3-16}
 &  & 3/3-0.3* & 91.94 & 0.97 & 2.88 & - & - & 0.85 & \textbf{4.70} & 1.75 & 1.38 & - & - & 0.17 & \textbf{3.29} \\ \hline
\end{tabular}
\label{tab:detail-cifar}
\vspace{-2ex}
\end{table*}

\textit{3) Analysis of Optimized Models.}
Table \ref{tab:detail-cifar} gives the performance results for the optimized models under a range of optimization parameters on CIFAR-10. To this end,
we have selected Jetson TX2 (a GPU-based edge device) and NCS2 (an ASIC-based edge device) as the two representative  platforms.
The total execution time for a neural network model consists of
the times spent on the system invocation, denoted as ``Sys.'', and
the forward computation of different operators, including ``Conv'' (convolution operators), ``Add'' (element-wise addition operators), ``BN'' (batch normalization operaters) and ``Others''.
Our results demonstrate the  performance improvements achieved from each component for the end-to-end inference.
First, the execution time of non-parametric operators such as ``Add" is non-negligible on edge platforms, indicating the necessity for optimizing non-parametric operators.
Second, our approach has successfully reduced the execution time of ``Add" with conservative pruning.
The performance of ``Conv" can be further improved by continued pruning.
For example, the optimized ResNet-32 model with ``\FuPruner-3/3'' decreases the execution time of ``Add'' from 1.24 ms to 0.00~ms on Jetson TX2, meaning that all the add operators in neural networks have been pruned away.
Furthermore, we can merge a BN operator into its previous convolution operator by adjusting and updating the weights, a common system-level optimization approach \cite{kang2018c}, with
the results of the BN-merged models denoted as ``*''.
There is neither an element-wise addition operator nor a batch normalization operator in the final pruned model, thereby enabling us to reduce significantly the execution time of neural network inference on edge intelligent devices. Overall,
our experimental results have confirmed the effectiveness of our approach for optimizing non-parametric operators on intelligent edge devices.

\subsection{Ablation Study}
To analyze each component of our proposed approach, ablation studies are conducted.

\begin{itemize}[leftmargin=2.5ex]

\begin{table}[t!]
	\centering
	\renewcommand\arraystretch{1.15}
	\caption{The performance of the fused models without filter pruning.}
	\begin{tabular}{|c||c|c|c|c|}
		\hline
		\multirow{2}{*}{\textbf{Model}} & \multicolumn{4}{c|}{\textbf{Inference Times (ms)}} \\ \cline{2-5}
		& \textbf{Original} & \textbf{1/3 Fused} & \textbf{2/3 Fused} & \textbf{3/3 Fused} \\ \hline
		ResNet-20 & 11.21 & 12.03 & 12.18 & 12.18 \\ \hline
		ResNet-32 & 16.97 & 18.34 & 18.73 & 18.89 \\ \hline
	\end{tabular}
	\label{tab:fused_ncs}
\end{table}

	\item[\textit{1)}] \textit{Operator Fusion without Pruning.}
	Table~\ref{tab:fused_ncs} shows the inference time of fused models without pruning on NCS with the original model used as a baseline.
	The fused model is more time-consuming than the original one due to the extra auxiliary operators fused into, illustrating the necessity of conservative pruning after fusion.

\item[\textit{2)}] \textit{Verifying Pruning Rates.}
	Figure~\ref{fig:pruning_rate} shows the model accuracy of different pruning rates for ResNet-20 and ResNet-32.
	The accuracy of the optimized model decreases with the increase of pruning rate and drops observably when the pruning rate is more than 30\%.
	Therefore, we set the filter pruning rate of \FuPruner from 0 to 0.3 in this paper.

\begin{figure}[t!]
	\centering
	\includegraphics[width=0.98\linewidth]{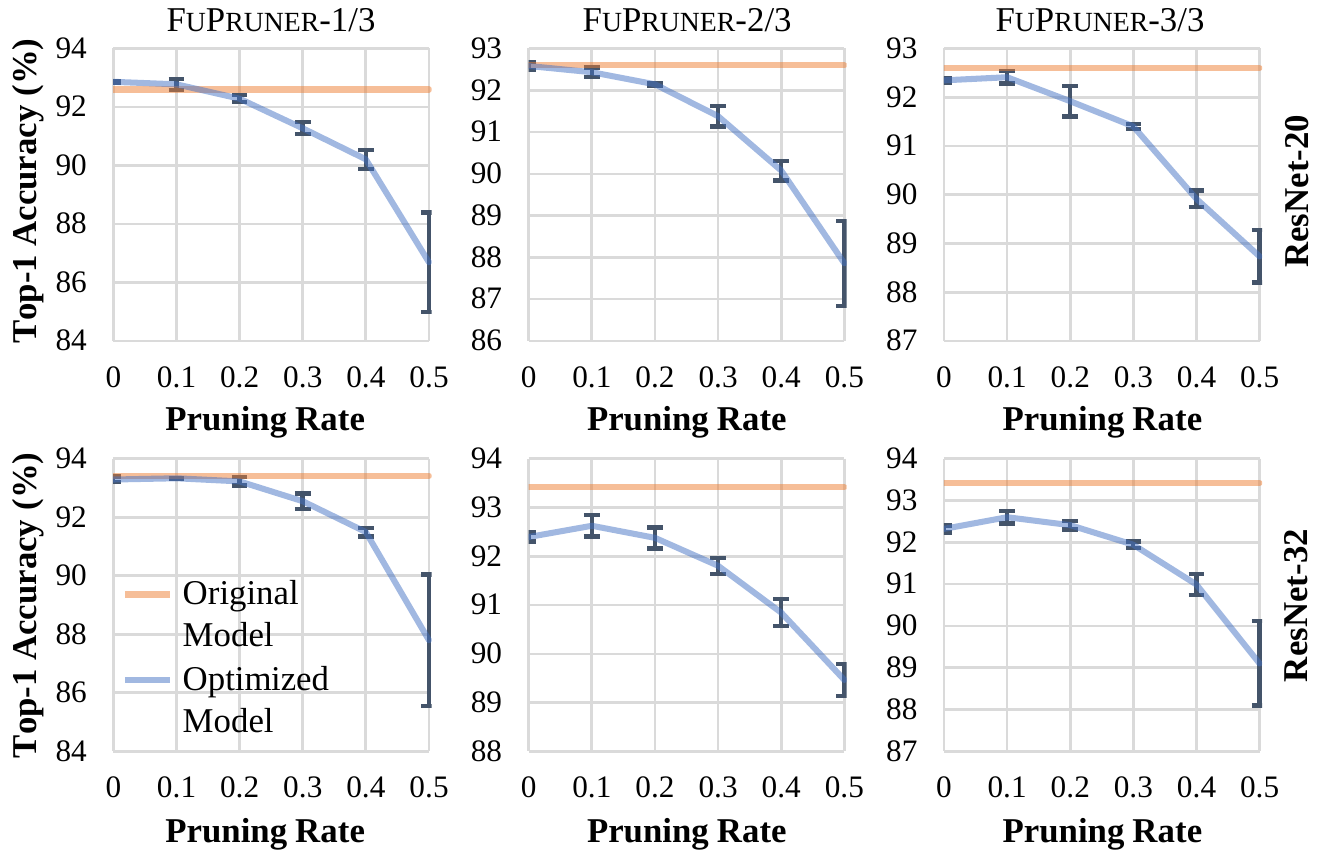}
	\caption{Model accuracy regarding different pruning rates. The orange line denotes the original model while the blue line denotes the optimized model. }
	\label{fig:pruning_rate}
\end{figure}

\item[\textit{3)}] \textit{Selection of the Fused Building Blocks.}
	We compare the accuracy of fusing different building blocks for ResNet-20, as shown in Table~\ref{tab:fusion_option}.
	We use ``(stage1, stage2, stage3)'' to denote the number of fused building blocks for each stage of the model.
	The accuracy may further improve if we carefully select the building blocks to be fused in \FuPruner based on hyper-parameter tuning.

\end{itemize}

\begin{table}[t!]
	\centering
	\renewcommand\arraystretch{1.15}
	\caption{The top-1 accuarcy of ResNet-20 with different fused blocks.}
    \begin{tabular}{|c|c||c|c|c|}
    \hline
    \multicolumn{2}{|c||}{\multirow{2}{*}{\textbf{\begin{tabular}[c]{@{}c@{}}Fused Building Blocks \\ in (stage1, stage2, stage3)\end{tabular}}}} & \multicolumn{3}{c|}{\textbf{Top-1 Accuracy (\%)}} \\ \cline{3-5}
    \multicolumn{2}{|c||}{} & $\bm{p_\varphi=0.1}$ & $\bm{p_\varphi=0.2}$ & $\bm{p_\varphi=0.3}$ \\ \hline
    \multirow{2}{*}{\begin{tabular}[c]{@{}c@{}}Fusing $3/9$\\ Building Blocks\end{tabular}} & (3,0,0) & 92.93 & 92.15 & 91.46 \\ \cline{2-5}
     & (1,1,1) & 92.40 & 92.46 & 91.83 \\ \hline
    \multirow{2}{*}{\begin{tabular}[c]{@{}c@{}}Fusing $6/9$\\ Building Blocks\end{tabular}} & (3,3,0) & 92.54 & 92.12 & 91.47 \\ \cline{2-5}
     & (2,2,2) & 93.04 & 92.30 & 91.84 \\ \hline
    \end{tabular}
	\label{tab:fusion_option}
\end{table}

\section{Discussion}

\noindent \textbf{Generality.}
This paper focuses on the convolutional neural network,
a widely used architecture that plays an important role in contemporary intelligent tasks.
The proposed auxiliary operators are currently  based on the computational pattern of convolution operations.
To extend our work to other scenarios such as optimizing recurrent neural networks, we need to analyze the performance characteristics of the target model on edge devices
and design a new set of auxiliary operators for the structures to be optimized.
Nevertheless, \FuPruner provides a new generic framework for optimizing non-parametric operators in the model compression approach  similarly  as for parametric operators, which can be generalized to support other structures and
 model acceleration domains.

\smallskip
\noindent \textbf{Applicability.}
\FuPruner represents a framework-indepen\-dent approach, making
it more generally applicable than other algorithm-level acceleration approaches, as it requires neither
 special implementations nor run-time system modifications.
Moreover, \FuPruner does not incur any extra run-time overhead,
making it well suited  for resource-constrained edge devices.
To determine that whether a DNN model on a target device can be accelerated by our approach or not, we can start with an analysis of the performance characteristics of the model on the device.
To achieve a realistic acceleration in the optimized model
by Equation~\ref{equ:sppedup},
 the time proportion of non-parametric operators in end-to-end inference must be non-negligible.
As revealed in our experimental results, the performance characteristics of neural networks on intelligent edge devices
are different from those on general-purposed platforms, providing more optimization opportunities for pruning non-parametric operators.
We envision for \FuPruner to become a useful acceleration technique for future edge intelligence.

\smallskip
\noindent \textbf{Soundness.}
The transformation of a computational graph is equivalence-preserving in the aggressive fusion method, without
causing any loss of accuracy,
as described in Section~\ref{sec:op_fusion}. However,
 the accuracy of a model may possibly decrease when applying our catalyzed-pruning approach to achieve both acceleration and compression. In this case,
\FuPruner provides configurable optimization options for controlling fusion and pruning, allowing  performance/accuracy trade-offs to be made.

\smallskip
\noindent\textbf{Limitations.}
Although \FuPruner shows the benefits of pruning parametric and non-parametric operators, there is room for further improvements.
First, we observe that pruning parametric or non-parametric operators has no significant optimization effect on some platforms, such as NCS, and pruning filters may even lead to performance degradation.
Therefore, in addition to focusing on the model compression itself, it is meaningful to design a mechanism to determine whether the compressed model has realistic acceleration.
Second, several hyper-parameters, such as fusion option and pruning rate, are necessary and predefined by the user in our approach so that a tuning process is needed currently.
We will explore how to automatically set the reasonable optimization hyper-parameters.
Finally, the accuracy of optimized models mainly depends on the filter pruning algorithm used and our aggressive operator fusion method can be combined with any filter pruning algorithm.
Therefore, using different pruning strategies for different fused models is another future research direction.

\section{Conclusion}
In this paper, we have introduced \FuPruner, a novel fusion-catalyzed pruning approach to optimize deep neural network models, which prunes parametric and non-parametric operators simultaneously for a faster inference on intelligent edge devices.
Evaluation with state-of-the-art residual neural networks on representative edge platforms has demonstrated the effectiveness of our approach.
We would like to apply \FuPruner to more complex intelligent applications, such as object detection, in the future work.

\bibliographystyle{IEEEtran}
\bibliography{ref}{}

\end{document}